\documentclass{article}

\usepackage{PRIMEarxiv}

\usepackage[utf8]{inputenc} 
\usepackage[T1]{fontenc}    
\usepackage{hyperref}       
\usepackage{url}            
\usepackage{booktabs}       
\usepackage{amsfonts}       
\usepackage{amsmath}
\usepackage{nicefrac}       
\usepackage{microtype}      
\usepackage{lipsum}
\usepackage{graphicx}
\usepackage{subcaption}
\usepackage{mathpazo}
\usepackage{sectsty}
\usepackage{tikz}
\usepackage[capitalize]{cleveref}
\usepackage{subcaption}
\usepackage[shortlabels,inline]{enumitem}
\usepackage{float}

\title{Analysis, forecasting and system identification of a floating offshore wind turbine using dynamic mode decomposition}

\author{
  Giorgio Palma$^{\star}$, Andrea Bardazzi, Alessia Lucarelli, Chiara Pilloton, Andrea Serani, \\ \textbf{Claudio Lugni, Matteo Diez}\\
  National Research Council-Institute of Marine Engineering, Rome, Italy\\
  $^\star$\texttt{giorgio.palma@cnr.it} \\
}

\begin{document}

\begin{tikzpicture}[remember picture,overlay]
   \node [rectangle, fill=cyan, fill opacity=0.5, anchor=north, minimum width=\paperwidth, minimum height=3cm] at (current page.north) {};

   \node [anchor=north, minimum width=\paperwidth, minimum height=3cm, text width=\textwidth, align=center, text height=5ex, text depth=15ex, align=left] at (current page.north) {
     \sffamily\small
     \textbf{This is a preprint submitted to:} \textit{Journal of Marine Science and Engineering}
   };
\end{tikzpicture}

\maketitle

\begin{abstract}
This article presents the data-driven equation-free modeling of the dynamics of a hexafloat floating offshore wind turbine based on the application of dynamic mode decomposition (DMD). 
All the analyses are performed on experimental data collected from an operating prototype.
The DMD has here used i) to extract knowledge from the dynamic system through its modal analysis, ii) for short-term forecasting from the knowledge of the immediate past of the system state, and iii) for the system identification and reduced order modeling.
The forecasting method for the motions, accelerations, and forces acting on the floating system is developed using Hankel-DMD, a methodological extension that includes time-delayed copies of the states in an augmented state vector.
The system identification task is performed by applying Hankel-DMD with control (Hankel-DMDc), which models the system including the effect of forcing terms. 
The influence of the main hyperparameters of the methods, namely the number of delayed copies in the state and input vector and the length of the observation time, is investigated with a full factorial analysis using three error metrics analyzing complementary aspects of the prediction: the normalized root mean square error, the normalized average minimum-maximum absolute error, and the Jensen-Shannon divergence.
A Bayesian extension of the Hankel-DMD and Hankel-DMDc is introduced by considering the hyperparameters as stochastic variables varying in suitable ranges defined after the full factorial analysis, enriching the predictions with uncertainty quantification.
Results show the capability of the approaches for short-term forecasting and system identification, suggesting their potential for real-time continuously-learning digital twinning and surrogate data-driven reduced order modeling.
\end{abstract}

\keywords{floating offshore wind turbine \and forecasting \and system identification \and dynamic mode decomposition \and DMD \and uncertainty quantification \and data driven \and equation free \and reduced order modeling \and ROM.}

\section{Introduction}\label{s:intro}
In the strive of containing the global temperature increase under 2$^\circ$C below pre-industrial levels, as set by the Paris Agreement \cite{falkner2016paris}, most countries have committed to reaching the goal of \textit{net zero emissions} by 2050, meaning that all the greenhouse gas emissions must be counterbalanced by an equal amount of removals from the atmosphere. 
To reach this critical and ambitious task for sustainable growth, the decarbonization of our society is a key aspect that passes through the decarbonization of energy production \cite{nastasi2022, IEA2021}.
The shift from fossil fuels to renewable sources for power production is to be considered the fundamental step in the process. 
Power generation is, in fact, responsible for 30\% of the global carbon dioxide emissions at the moment.
In 2018, the European Union set intermediate targets of 20\%  of energy obtained from renewable resources by 2020 and 32\% by 2030, the latter has been raised to 42.5\% (with the aspiration of reaching 45\%) by amending the Renewable Energy Directive in 2021 \cite{EURenewableEnergyDirective}.
Reaching the mentioned targets means almost doubling the existing share of renewable energy in the EU.  

Wind energy technology has been identified as one of the most promising ones, along with photovoltaic, for power production from renewable sources. 
Several growth scenarios predict a prominent role of wind power, exceeding the 35\% share of the total electricity demand by 2050 \cite{prakash2019future},
representing a nearly nine-fold rise in the wind power share in the total generation mix compared to 2016 levels.
Offshore wind energy production has a bigger growth potential compared to its onshore counterpart. 
The reasons are connected to fewer technical, logistic, and social restrictions of the former.
Offshore installed turbines may exploit abundant and more consistent winds, helped by the reduced friction of the sea surface and the absence of surrounding hills and buildings \cite{Konstantinidis2016}. 
In addition, offshore wind farms benefit a greater social acceptance, a minor value of their occupied space, and the possibility of installing larger turbines with fewer transportation issues than onshore \cite{pustina2020, pustina2023, lopez2022}.
The 2023 Global Offshore Wind Report predicts the installation of more than 380 GW of offshore wind capacity worldwide in the next ten years \cite{council2020global}.

The exponential growth of the sector passes through the possibility of realizing floating offshore plants, enabling the exploitation of sea areas with deeper water that make fixed-foundation turbines not a feasible/affordable solution (indicatively deeper than 60 m).
The main advantage of exploiting deep offshore sea areas relies on the abundant and steady winds characterizing them.
One of the main limiting factors in the reduction of the levelized cost of energy (LCOE) of advanced floating offshore wind turbines (FOWTs) is the current size and cost of their platforms. 
Its reduction, alongside the development of advanced moorings, improved control systems, and maintenance procedures is among the most impacting technical goals research activities are focusing on. 

The power production by FOWTs presents additional challenges compared to the fixed wind turbine counterpart (on- or offshore) which are inherent to their floating characteristic that adds six degrees of freedom to the structures. 
Nonlinear hydrodynamic loads, wave-current interactions, aero-hydrodynamics coupling producing negative aerodynamic damping, and wind-induced low-frequency rotations are among the main causes of large amplitude motions of the platform. These are in turn causes of reductions in the average power output of the power plant and increases in the fluctuations of the produced power. 
Both quantity and quality of the power production are affected and the structure and all the components (blades, cables, bearings, etc.) also suffer increased fatigue-induced wear from non-constant loadings \cite{mcmorland2022}
(about 20\% of operation and maintenance costs come from blade failures \cite{Florian2015} and almost 70\% of the gearbox downtime is due to bearing faults \cite{deazevedo2016}).

The floating wind turbine operations are considerably altered by the stochastic nature of wind, waves, and currents in the sea environment, which excite platform motion leading to uncertainties in structural loads and power extraction capability.
As shown first in \cite{jonkman2007}, waves are responsible for a large part of the dynamic excitation of a FOWT. Rotor speed fluctuations are 60\% larger when the same turbine is operated in a floating environment compared to onshore installation, and the difference has been shown to increase with increasing wave conditions.
Therefore, it is essential to develop appropriate strategies to improve the FOWTs' platform stability and maximize the turbines' energy conversion rate, achieving better LCOE with higher power production and lower maintenance operation costs. 

Both passive and active technologies have been studied and developed to the scope, such as tuned mass dampers \cite{Si2014, Li2016, Dong2020} mounted on different floaters, ballasted buoyancy cans \cite{Edem2022, Galvan2018, Grant2023}, gyro-stabilizers \cite{Palraj2021}, blade pitch and/or torque controllers \cite{lopez2022, Salic2019, Tiwari2016, pustina2020, pustina2023}. 
Developing advanced control systems is a high-potential cost reduction strategy for offshore wind turbines impacting at multiple levels: effective control strategies may increase the energy production which has a direct impact on LCOE; a reduction of the platform motions may help in reducing their sizes and costs; reduced vibratory loads on the turbine's components helps increasing their lifetime and reducing maintenance costs. 
A thorough comparison of various controllers designed for managing vibratory loads is provided in \cite{awada2021}. 

Both feedback \cite{Larsen2007,jonkman2008,Yu2018, bakka2012, Betti2012, Lemmer2020, pustina2020, Sarkar2021, pustina2023} and feedforward \cite{Harris2006,laks2010,dunne2010,dunne2011,dunne2012,Raach2014,Schlipf2015,ma2018,Simley2020,fontanella2021} control systems have been successfully developed for the effective control of FOWT.
The two philosophies may also be effectively coupled, creating a feedforward control with a feedback loop such as in \cite{Al2020}, where real-time prediction of the free-surface elevation is exploited to compensate for the wave disturbances on the FOWT.

Feedback algorithms may take advantage of accurate models of the controlled systems for improved performance, at the same time feedforward controllers rely on forecasting techniques to estimate upcoming disturbances or changes in the state to be controlled. 
Hence, the predictive algorithm plays a primary role in the success of the control strategy.

Although white and grey-box models for the FOWT dynamics can be obtained \cite{Otter2022,Hoeg2023}, the modeling process can be extremely complex and clearly representing the operational status of the turbine and platform in random meteorological conditions is not trivial.
Recently, data-driven methods demonstrated to be a powerful alternative for the identification of dynamic systems and the forecasting of their response. 
In particular, advanced machine learning and deep learning algorithms have been successfully applied to predict FOWT motions and loads.
Several examples can be found in the literature.
\cite{wang2023} used a multi-layer feed-forward neural network to predict maximum blade and tower loadings and maximum mooring line tension, using wind speed, turbulence intensity, meaningful wave height, and spectral peak period as the model’s input parameters.
\cite{zhang2022} built a data-driven prediction model for the FOWT output power, the platform pitch angle, and the blades' flapwise moment at the root using wind speed, wave height, and blade pitch control variables as inputs by training a gated recurrent neural network (GRNN).
The study in \cite{barooni2024} adopts a convolutional neural network merged with a GRNN forecasting the dynamic behavior of FOWTs.
Long short-term memory (LSTM) networks are the subject of the studies in \cite{grafe2024}, where fairlead tension, surge, and pitch motions are predicted with a data-driven method including onboard sensors measurements and lidar inflow data.
Self-attention method is integrated with LSTM in \cite{deng2024} to improve accuracy in predicting the motion response of a FOWT in wind-wave coupled environments. 
The hybridization of LSTM with empirical mode decomposition (EMD) is studied in \cite{ye2022}, where the neural network is used to predict the subcomponents of the EMD process for the short-term prediction of motions of a semi-submersible platform, and in \cite{song2025}, studying FOWT motion response prediction under different sea states.

Albeit powerful, machine learning and deep learning methods typically require large training datasets, comprising a range of operating conditions as complete as possible to learn patterns that generalize to \textit{new} situations. In addition, the training of such algorithms can be computationally expensive, and typically not compatible with real-time learning and digital twinning, where the system characteristics and response to external perturbations change with the system aging.

Dynamic mode decomposition offers an interesting alternative for data-driven and equation-free modeling \cite{schmid2010,kutz2016dynamic,mezic2021koopman}. 
The method is based on the Koopman operator theory, an alternative formulation of dynamical systems theory that provides a versatile framework for the data-driven study of high-dimensional nonlinear systems \cite{mezic2017}. 

DMD builds a reduced-order linear model of a dynamical system, approximating the Koopman operator. The approach requires no specific knowledge and assumption about the system's dynamics and can be applied to both empirical and simulated data.
The model is obtained with a direct procedure from a small set of multidimensional input-output pairs, constituting, from a machine learning perspective, the training phase.  

The data-driven nature, non-iterative training process, and data efficiency of DMD have contributed to its widespread adoption as a reduced-order modeling technique 
and real-time forecasting tool
in various fields. These include fluid dynamics and aeroacoustics \cite{rowley2009, schmid2010, Tang2012, Semeraro2012, Song2013}, epidemiology \cite{Proctor2015}, neuroscience \cite{brunton2016}, finance \cite{mann2016}, among others.

The literature features several methodological extensions of the original DMD algorithm aimed at improving the accuracy of the decomposition and, more generally, broadening the method's capabilities. 
Particularly relevant to this study are the Hankel-DMD \cite{mezic2017,Brunton2017,kamb2020time,mezic2021koopman,serani2023}, the DMD with control (DMDc) \cite{proctor2016dynamic, Proctor2018}, and their combination in the Hankel-DMD with control (Hankel-DMDc) \cite{Brunton2021,ZAWACKI2023}. 

The Hankel-DMD, also referred to as Augmented-DMD \cite{serani2023}, Time-Delay Coordinates Extended DMD \cite{Brunton2021}, and Time Delay DMD \cite{Dylewsky2022}, has proven to be a powerful tool for enhancing the linear model's ability to capture significant features of nonlinear and chaotic dynamics.
This is achieved by extending the system state with time-delayed copies, which, in the limit of infinite-time observations, yields the true Koopman eigenfunctions and eigenvalues \cite{mezic2017}.
For instance, it has been effectively used in \cite{mohan2018data} and \cite{Dylewsky2022} to predict the short-term evolution of electric loads on the grid.

The DMDc extends the DMD framework to handle externally forced systems, allowing for the separation of the system's free response from the effects of external inputs. 
Notable applications include \cite{ALJIBOORY2024}, which developed a novel real-time control technique for unmanned aerial quadrotors using DMDc enabling the control system to adapt promptly to environmental or system behavior change.
Another example is \cite{dawson2015}, where DMDc has been applied to simulation data to create a reduced-order model (ROM) of the forces acting on a rapidly pitching airfoil.

In \cite{Brunton2021} the algorithmic variant combining control and state augmentation with time-delayed copies is called Time-Delay Coordinates DMDc, and it is introduced using the same number of delayed copies for both the state and the input. 
Similarly, \cite{ZAWACKI2023} introduced the Dynamic Mode Decomposition with Input-Delayed Control, which, as the name suggests, includes time-delayed copies of the inputs only.

Several studies have demonstrated the effectiveness of DMD in forecasting complex system behaviors in the marine environment, such as \cite{diez2022datadriven} where DMD is applied for the forecasting of ship trajectories motions and forces. Additionally, \cite{serani2023} conducted a statistical evaluation of DMD's predictive performance, incorporating state augmentation techniques such as derivatives and time-delayed copies of the state.
\cite{Diez2024} provides a comparative analysis for the same naval application of DMD-based prediction algorithms and various neural network architectures, including standard and bidirectional long-short-term memory networks, gated recurrent units, and feedforward neural networks. 
Furthermore, \cite{diez2022snh} studied the hybridization of DMD with artificial neural networks to enhance prediction accuracy. 

The objective of this paper is to explore, for the first time to the best of the authors' knowledge, the use of DMD and its variants to extract knowledge, develop a forecasting algorithm, and a system identification method for predicting relevant quantities of a FOWT from experimental data. 
In particular, Hankel-DMD is used as a data-lean forecasting method, producing short-term forecasting (nowcasting) from the immediate past history of the system state, with a continuously learning data-driven reduced order model, suitable for digital twinning and real-time predictions.
On the other side, Hankel-DMDc is applied as an effective and efficient approach to model-free system identification, aiming to create an accurate ROM for the long-term prediction of the platform motions and loads from the knowledge of the wave elevation in the proximity of the platform and the wind speed.
In this work, the nowcasting and system identification tasks are performed on experimentally measured data, however, the Hankel-DMD and Hankel-DMDc methods here developed directly apply also to different data sources such as simulations of various fidelity levels.
The effect on the predictions of the main hyperparameters of the methods is studied with a full-factorial design of experiment, assessing the performances using three error metrics and identifying the most promising configurations. 

In addition, novel Bayesian extensions of Hankel-DMD and Hankel-DMDc are introduced to include uncertainty quantification in the methods' predictions by considering their hyperparameters as stochastic variables. 
The stochastic hyperparameter variation ranges are identified after the deterministic analyses, and the results from the deterministic and Bayesian methods are compared using the same test sequences.

The methods are applied to real-life measured data obtained by various sensors mounted on a scale prototype of a 5MW Hexafloat FOWT first of its type.
The experimental activity has been conducted as part of the National Research Project {\em RdS-Electrical Energy from the Sea}, funded by the Italian Ministry for the Environment (MaSE) and coordinated by CNR-INM. 

The efficiency of the DMD-based methods is provided by the small dimensionality of the relevant state variables and the low computational cost required for both the model construction (training) and exploitation (prediction), as opposed to more data and resource-intensive machine learning methods.

The paper is organized as follows. 
\Cref{s:matmeth} presents the wind turbine test case, details the DMD methods applied, and introduces the performance metrics used to assess the predictive performances of the algorithms.
The numerical setup and the data preprocessing are described in \cref{s:numset}. 
\Cref{s:res} collects the results from the modal analysis and the forecasting of the quantities of interest obtained with the deterministic Hankel-DMD and its Bayesian version.
Finally, conclusions about the conducted analyses are resumed in \cref{s:conc}.

\section{Material and methods}\label{s:matmeth}
\subsection{Wind turbine test case}
\begin{figure}[ht]
  \centering
  \captionsetup[subfigure]{justification=centering}
  \begin{subfigure}[b]{0.56\linewidth}
    \includegraphics[width=\linewidth]{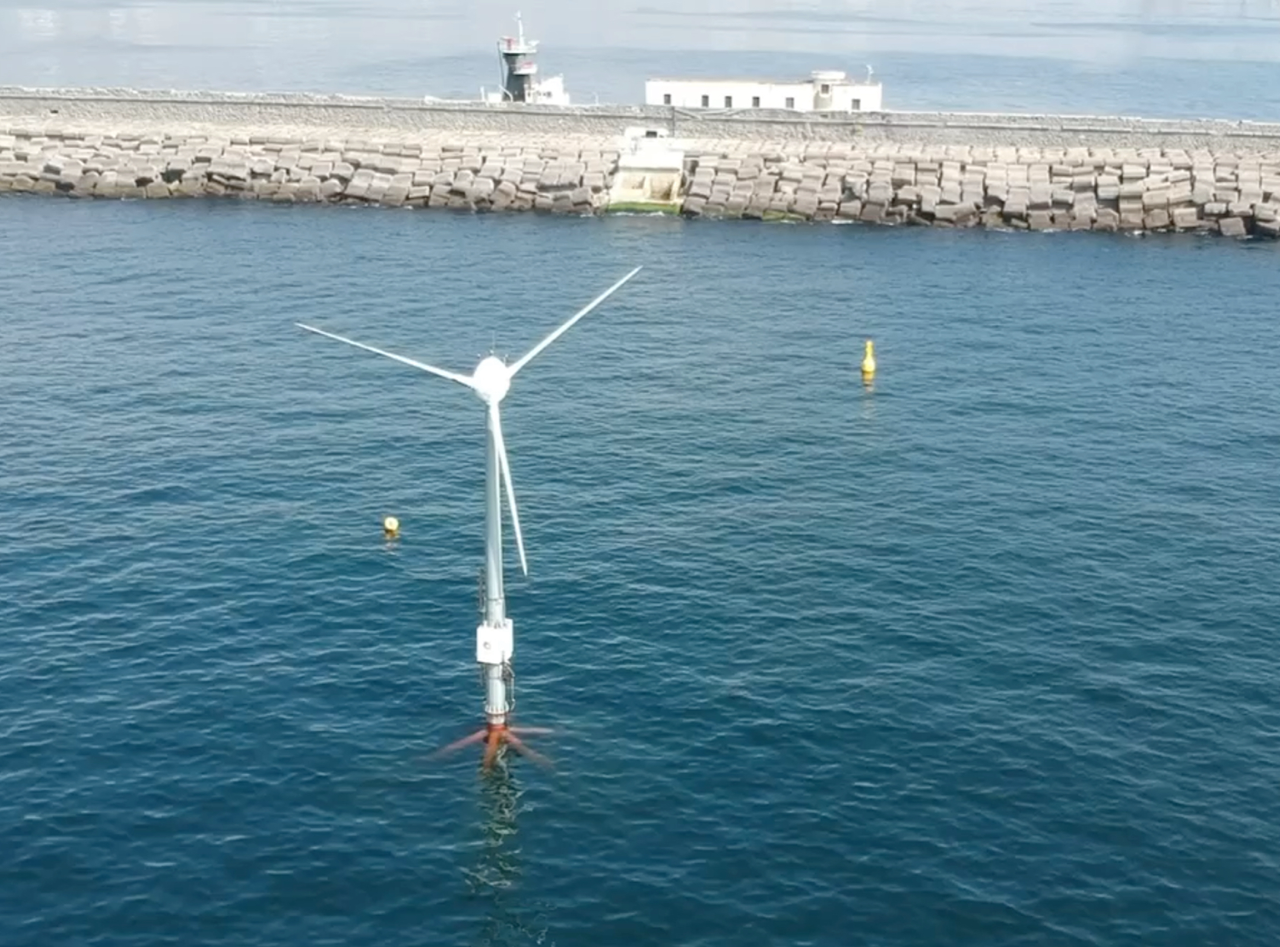}
    \caption{} \label{fig:setup1}  
  \end{subfigure}  
  \begin{subfigure}[b]{0.26\linewidth}
    \includegraphics[width=\linewidth]{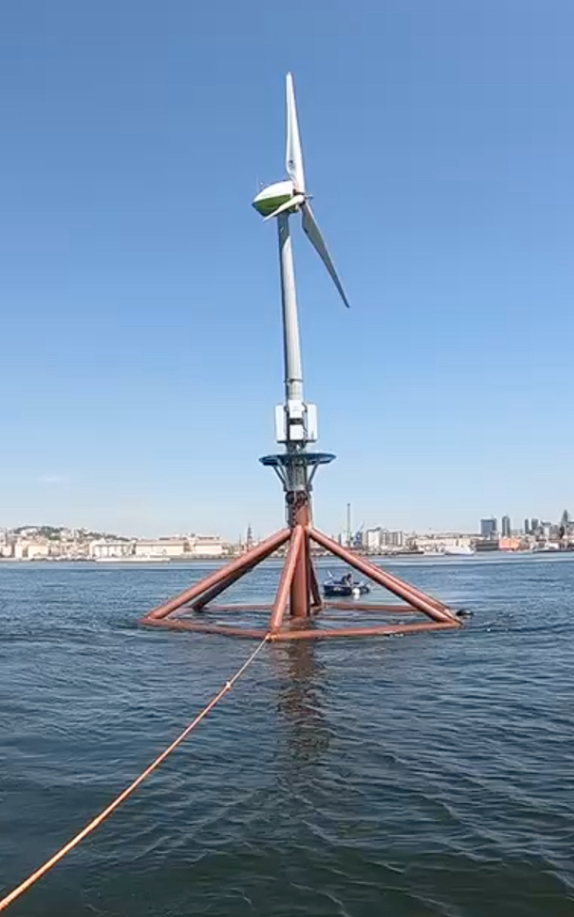}
    \caption{}\label{fig:setup2}
  \end{subfigure}
  \caption{(\subref{fig:setup1}) Aerial view of In-MaRELab with the Hexafloat FOWT during the tests at sea in 2021. (\subref{fig:setup2}) View of the Hexafloat FOWT during the towing stage from the shipyard to the test site in 2024.}\label{fig:setup}
\end{figure}

The presented analyses are conducted on a set of experimental data collected on the prototype of a FOWT built and tested at the Interdisciplinary Marine Renewable Energy sea Lab (In-MaRELab), see Figure~\ref{fig:setup1},            
offshore the Naples port, right in front of the breakwater Molo San Vincenzo. 
The prototype and the experimental activity at sea are part of the National Research Project {\em RdS-Electrical Energy from the Sea}, funded by the Italian Ministry for the Environment (MaSE) and coordinated by CNR-INM. 
The FOWT is a 1:6.8 scale prototype of a 5MW FOWT, the first one 
existing at sea for the Hexafloat concept. The floater is a Saipem patented lightweight semisub platform consisting of a hexagonal tubular steel structure around a central column, and a deeper counterweight, connected to the floater by six tendons (one for each corner of the hexagon) in synthetic material.                       
The floater hosts a Tozzi Nord TN535 10-kW wind turbine \cite{TN535}, originally designed for onshore application and, within the present RP, suitably modified on the               
electrical part for the specific aims of the offshore application (see Figure~\ref{fig:setup2}).                    
The FOWT is anchored with three drag anchors located at $30$ m of water depth through three mooring lines in catenary configuration at a relative angle of $120$ degrees counterclockwise, with $M_1$ directed towards the breakwater and orthogonal to it. 

The studied dataset includes 12-hour synchronized time histories of
\begin{enumerate}
   \item the loads applied to one of the three moorings of the platform ($M_3$), and three of the six tendons connecting the counterweight to the floater ($T_1$, $T_5$, and $T_6$), as measured by a system of underwater load cells LCM5404 with work limit load (WLL) of $2.3$ tons and $4.5$ tons, respectively;  
   \item the acceleration along three coordinate axes ($\dot{u}$, $\dot{v}$, $\dot{w}$), the pitch and roll angles ($\theta$, $\phi$), and the respective angular rates ($\dot{\theta}$, $\dot{\phi}$ ), collected by the Norwegian Subsea MRU 3000 inertial motion unit;            
   \item the power extracted by the wind turbine ($P$), estimated by a PLC through a direct measure of the electrical quantities at the generator on board the nacelle of the wind turbine, the rotor angular velocity ($\Omega$) measured by two sensors in continuous cross-check, the relative wind speed ($V_w$) through two different anemometers positioned on the nacelle, behind the rotor. All signals were collected by the PLC on the nacelle with a variable but well-known sample frequency of approximately $1$ Hz;
   \item the wave elevation ($h_w$) measured by a pressure transducer integrated into the Acoustic Doppler Current Profiler (ADCP) Teledyne Marine-Sentinel V20, located at a distance of approximately $50$ m from the FOWT in the SE direction.
\end{enumerate}
The state of the system is hence composed as $\mathbf{x} = \{$$T_1$, $T_5$, $T_6$, $M_3$, $\phi$, $\theta$, $\dot{\phi}$, $\dot{\theta}$, $\dot{u}$, $\dot{v}$, $\dot{w}$, $P$, $\Omega$, $V_w$, $h_w$$\}$.

The data consider a window of continuous operations in extreme weather conditions.
The system may show induced strongly non-linear physical dynamics in such weather conditions. In particular, the intense wind causes a nonlinear behavior in the extracted power and blades' rotating speed, showing a saturation to the maximum values supported by the machine. This poses the DMD-based methods in a challenging condition for modal analysis and prediction.
An average incoming wave period is identified from the peak in the wave elevation spectrum and used in the following as reference period $\hat T=7.3143$ $s$ (${\hat f} = 0.1367$ Hz). 

\subsection{Dynamic mode decomposition}\label{s:dmd}
Dynamic mode decomposition was originally presented in \cite{schmid2008dynamic} and \cite{schmid2010} to identify spatiotemporal coherent structures from high-dimensional time series data, providing a linear reduced-order representation of possibly nonlinear system dynamics.
Given a time series of data, the DMD computes a set of modes with their associated frequencies and decay/growth rates \cite{Brunton2021}. 
When the analyzed system is linear, the modes obtained by the DMD correspond to the system's linear normal modes. 
The potential of the DMD in the analysis of nonlinear systems comes from its close relation to the spectral analysis of the Koopman operator \cite{rowley2009}.
The Koopman operator theory is built upon the original work in \cite{Koopman1931}, defining the possibility of transforming a non-linear dynamical system into a possibly infinite-dimensional linear system \cite{Tu2014,Proctor2018}. 
DMD is an equation-free data-driven approach that was shown by \cite{rowley2009} to be a computation of the Koopman operator for linear observables \cite{Marusic2024}.
Approximating the eigenmodes and eigenvalues of the infinite-dimensional linear Koopman operator, the DMD analysis thus provides a linearized finite-dimensional representation of the dynamics of the nonlinear original system \cite{kutz2016dynamic}.

The DMD algorithm has been generalized and improved since its introduction. Its state-of-the-art definition has been given by \cite{Tu2014} and is resumed in the following.
Consider a dynamical system described by
\begin{equation}\label{eq:nlsys}
    \frac{\mathrm{d}\mathbf{x}}{\mathrm{dt}} = \mathbf{f}\left( \mathbf{x}, t, \gamma \right)
\end{equation}
where $\mathbf{x}(t) \in \mathcal{R}^n$ represents the system's state at time $t$, $\gamma$ contains the parameters of the system, and $\mathbf{f}(\cdot)$ represents its dynamics. In typical applications, the state $\mathbf{x}$ may represent the discretization of a partial differential equation at several spatial points, or be composed by multiple-variable time series of a dynamical system, and thus is generally large with $n\gg 1$.
Approximating \cref{eq:nlsys} with DMD leads to a locally linear dynamical system defined as
\begin{equation}\label{eq:dmdsys}
    \frac{\mathrm{d}\mathbf{x}}{\mathrm{dt}} = \mathcal{A}\mathbf{x}
\end{equation}
which solution can be expressed in terms of eigenvalues $\omega_k$ and eigenvectors $\phi_k$ of the matrix $\mathcal{A}$
\begin{equation}\label{eq:soleigen}
    \mathbf{x}(t) = \sum_{k=1}^n \phi_k q_k(t) = \sum_{k=1}^n \phi_k  b_k \text{exp}(\omega_k t),
\end{equation}
where the coefficients $b_k$ are the coordinates of the initial condition $\mathbf{x}_0$ in the eigenvector basis, $\mathbf{b}=\boldsymbol{\Phi}^{-1} \mathbf{x}_0$

In practical applications, the state of the system is measured at $m$ discrete time steps $t_j=j\,\Delta t$, and can be expressed as $\mathbf{x}_j = \mathbf{x}(j\Delta t)$ with $j = 1,\dotsc,\,m$. This data is governed by a discrete-time dynamical system, equivalent to the one of \cref{eq:nlsys}
\begin{equation}\label{eq:dtsys}
    \mathbf{x}_{j+1} = \mathrm{\mathbf{F}}(\mathbf{x}_j,j,\gamma)
\end{equation}
Equivalently, the DMD approximation can be written as
\begin{equation}\label{eq:dmdsysdisc}
    \mathbf{x}_{j+1} = \mathrm{\mathbf{A}}\mathbf{x}_j, \hspace{1cm} \text{with} \hspace{1cm} \mathrm{\mathbf{A}}=\text{exp}(\mathcal{A}\Delta t)
\end{equation}
For each time step $j$, a snapshot of the system is defined as the column vector collecting the measured full state of the system $\mathbf{x}_j$. 
Two matrices can be obtained by arranging the available snapshots as follows
\begin{equation}\label{eq:datamatrices}
    \mathrm{\mathbf{X}} = \left[ \mathbf{x}_{1} \quad \mathbf{x}_{2} \quad \dotsc \quad \mathbf{x}_{m-1} \right],  \hspace{1cm} \mathrm{\mathbf{X}}' = \left[ \mathbf{x}_{2} \quad \mathbf{x}_{3} \quad \dotsc \quad \mathbf{x}_{m} \right]
\end{equation}
such that \cref{eq:dmdsysdisc} may be written in terms of these data matrices
\begin{equation}
    \mathrm{\mathbf{X}}' \approx \mathrm{\mathbf{A}} \mathrm{\mathbf{X}} 
\end{equation}
Hence, the matrix $\mathrm{\mathbf{A}}$ can be constructed using the following approximation
\begin{equation}
    \mathrm{\mathbf{A}} \approx \mathrm{\mathbf{X}}' \mathrm{\mathbf{X}}^{\dagger} 
\end{equation}
where $\mathrm{\mathbf{X}}^{\dagger}$ is the Moore-Penrose pseudoinverse ox $\mathrm{\mathbf{X}}$, which minimize $||\mathrm{\mathbf{X}}' - \mathrm{\mathbf{AX}}||_F$, where $||\cdot ||_F$ is the Frobenius norm.
The pseudoinverse of $\mathrm{\mathbf{X}}$ can be efficiently evaluated using the singular value decomposition (SVD) as $\mathrm{\mathbf{X}}^{\dagger} = \mathrm{\mathbf{V}} \boldsymbol{\Sigma}^{-1} \mathrm{\mathbf{U}}^*$, where $^*$ denotes the complex conjugate transpose. 
The matrix $\mathrm{\mathbf{\tilde A}}$ is evaluated projecting $\mathrm{\mathbf{A}}$ onto the POD modes in $\mathrm{\mathbf{U}}$
\begin{equation}
    \mathrm{\mathbf{\tilde A}} = \mathrm{\mathbf{U}}^* \mathrm{\mathbf{A}} \mathrm{\mathbf{U}},
\end{equation}
and its spectral decomposition can be evaluated
\begin{equation}
    \mathrm{\mathbf{\tilde A}} \mathrm{\mathbf{W}} = \mathrm{\mathbf{W}} \mathrm{\mathbf{\Lambda}}
\end{equation}
The diagonal matrix $\mathrm{\mathbf{\Lambda}}$ contains the DMD eigenvalues $\lambda_k$, while the DMD modes $\phi_k$ constituting the matrix $\mathrm{\mathbf{\Phi}}$ are then reconstructed using the eigenvectors $\mathrm{\mathbf{W}}$ and the time-shifted data matrix $\mathrm{\mathbf{X}}'$ (this algorithm is referred to as the exact-DMD \cite{Tu2014})
\begin{equation}\label{eq:dmdmodes}
    \mathrm{\mathbf{\Phi}} = \mathrm{\mathbf{X}}' \mathrm{\mathbf{V}} \mathrm{\mathbf{\Sigma}}^{-1} \mathrm{\mathbf{W}}
\end{equation}
The state-variable evolution in time can be approximated by the modal expansion of \cref{eq:soleigen}, where $\omega_k = \ln{(\lambda_k)}/\Delta t$, starting from an initial condition corresponding to the end of the measured data $\mathrm{\mathbf{b}}=\mathrm{\mathbf{\Phi}}^{-1} \mathrm{\mathbf{x}}_m$.

\subsection{Dynamic mode decomposition with control}\label{s:dmdc}
The original DMD characterizes naturally evolving dynamical systems. The DMDc aims at extending the algorithm to include in the analysis the influence of forcing inputs and disambiguate it from the unforced dynamics of the system \cite{proctor2016dynamic}. 
While DMD (and its Hankel extension) is considered in this work for the nowcasting task, DMDc is considered more suitable for the system identification task obtaining a reduced-order model for the prediction of FOWT platform motions and loads acting on moorings and tendons, forced by two control variables, namely, wind velocity and wave elevation.

The DMDc formulation can be obtained from a forced dynamic system in its discrete-time version:
\begin{equation} \label{eq:discdynsys}
    \mathbf{x}_{j+1} = \mathrm{\mathbf{F}}(\mathbf{x}_j,j,\gamma,\mathbf{u_j}),
\end{equation}
similarly to \cref{eq:dtsys} but including the forcing input $\mathbf{u}  \in \mathbb{R}^l$ in the nonlinear mapping describing the evolution of the state.
DMDc modeling approximates \cref{eq:discdynsys} as:
\begin{equation}\label{eq:dmdcdsys}
    \mathbf{x}_{j+1} = \mathrm{\mathbf{A}}\mathbf{x}_j + \mathbf{B}\mathbf{u}_j
\end{equation}
where $\mathbf{B} \in \mathbb{R}^{n \times l}$, and the discrete-time matrix $\mathbf{A} \in \mathbb{R}^{n\times n}$.

Introducing the vector $\mathbf{y}_j$
\begin{equation}\label{eq:Y}
\mathbf{y}_j=
\begin{bmatrix}
\mathbf{x}_j \\
\mathbf{u}_j\\
\end{bmatrix},
\end{equation}
\cref{eq:dmdcdsys} can be rewritten in a form close to \cref{eq:dtsys}:
\begin{equation}\label{eq:dmdSIY}
\mathbf{x}_{j+1}=\mathbf{Gy}_j, \hspace{1cm} \text{with} \hspace{1cm} \mathbf{G}=
\begin{bmatrix}
    \mathbf{A} & \mathbf{B} \\
\end{bmatrix}
\end{equation}.
The collected data are arranged in the following matrices:
\begin{equation}\nonumber
\mathbf{Y}=
\begin{bmatrix}
\mathbf{y}_j & \mathbf{y}_{j+1} & \dots & \mathbf{y}_{m-1}\\
\end{bmatrix},
\end{equation}
\begin{equation}\label{eq:XX'}
\mathbf{X}'=
\begin{bmatrix}
\mathbf{x}_{j+1} & \mathbf{x}_{j+2} & \dots & \mathbf{x}_{m}\\
\end{bmatrix},
\end{equation}
and the DMD approximation of the matrix $\mathbf{G}$ can be obtained from
\begin{equation}\label{eq:approxG}
\mathbf{G}\approx\mathbf{X}'\mathbf{Y}^{\dag},
\end{equation}
where $\mathbf{Y}^{\dag}$ is the Moore-Penrose pseudo-inverse of $\mathbf{Y}$, which minimizes $\|\mathbf{X}'-\mathbf{GY}\|_F$, where $\|\cdot\|_F$ is the Frobenius norm. 
In this way, the matrices $\mathbf{A}$ and $\mathbf{B}$ are obtained as the ones providing the best fitting of the sampled data in the least squares sense.
Again, the pseudoinverse of $\mathbf{Y}$ can be efficiently evaluated using the singular value decomposition (SVD)
\begin{equation}\label{eq:gsvd}
    \mathbf{G} = \mathbf{X}' \mathbf{V} \boldsymbol{\Sigma}^{-1}\mathbf{U}^*.
\end{equation}
The DMD approximations of the matrices $\mathbf{A}$ and $\mathbf{B}$ are obtained by splitting the operator $\mathbf{U}$ in $\mathbf{U_1} \in \mathbb{R}^{n \times n}$ and $\mathbf{U_2} \in \mathbb{R}^{l\times n}$ \footnote{Due to the low dimensionality of data in the current context, the pseudoinverse is computed using the full SVD decomposition with no rank truncation. Otherwise, truncating the SVD to rank $p$ one would have $\mathbf{U_1} \in \mathbb{R}^{n\times p}$ and $\mathbf{U_2} \in \mathbb{R}^{l\times p}$.}. 
\begin{equation}\label{eq:abdmdc}
    \mathbf{A} = \mathbf{X'}\mathbf{V}\boldsymbol{\Sigma}^{-1}\mathbf{U_1^*}, \hspace{1cm} \mathbf{B} = \mathbf{X'}\mathbf{V}\boldsymbol{\Sigma}^{-1}\mathbf{U^*_2}.
\end{equation}

\Cref{eq:dmdcdsys} shows how the system dynamics can be predicted once the DMDc approximations of $\mathbf{A}$ and $\mathbf{B}$ are retrieved, with given initial conditions and the sequence of inputs $\mathbf{u}_j$.

\subsection{Hankel extension to DMD and DMDc}\label{s:hdmd}
The standard DMD and DMDc formulations approximate the Koopman operator in a restricted space of linear measurements, creating a best-fit linear model linking sequential data snapshots \cite{schmid2010,kutz2016dynamic}. 
The linear DMD provides a locally linear representation of the dynamics that can't capture many essential features of nonlinear systems.
The augmentation of the system state is thus the subject of several DMD algorithmic variants \cite{otto2019,takeishi2017,Lusch2018,Brunton2021}, aiming to find a coordinate system (or \textit{embedding}) that spans a Koopman-invariant subspace, to search for an approximation of the Koopman operator valid also far from fixed points and periodic orbits in a larger space.
The need for state augmentation through additional observables is even more critical for applications in which the number of states in the system is small, typically smaller than the number of available snapshots.
However, there is no general rule for defining these observables and guaranteeing they will form a closed subspace under the Koopman operator \cite{brunton2016b}.

The Hankel-DMD \cite{mezic2017} is a specific version of the DMD algorithm that has been developed to deal with the cases of nonlinear systems in which only partial observations are available such that there are \textit{latent} variables \cite{Brunton2021}.
The state vector is thus augmented, embedding $s$ time-delayed copies of the original variables. 
The Hankel-DMD with control (Hankel-DMDc) involves, in addition, the augmentation of the input vector with $z$ time-delayed copies of the original forcing variables. 
This results in an intrinsic coordinate system that forms an invariant subspace of the Koopman operator (the time-delays form a set of observable functions that span a finite-dimensional subspace of Hilbert space, in which the Koopman operator preserves the structure of the system \cite{Brunton2017, Pan2020}).
The Hankel augmented DMD variants, hence, can better represent the underlying dynamics of the systems allowing the capture of their important nonlinear features.

The formulation of the Hankel-DMD can be obtained starting from the DMD presented in \cref{s:dmd}. 
The dynamical system is approximated by Hankel-DMD as:
\begin{equation}\label{eq:hdmdsysdisc}
    \mathbf{\hat x}_{j+1} = \mathrm{\mathbf{A}}\mathbf{\hat x}_j, \hspace{1cm}
\end{equation}
where the augmented state vector is defined as $\hat{\mathbf{x}}_j = [\mathbf{x}_j,\,\mathbf{x}_{j-1},\,\dots\,,\mathbf{x}_{j-s}]^T$. 

The only modification in the algorithm for the eigenvalue and eigenvector identification described in Eqs. \ref{eq:datamatrices}--\ref{eq:dmdmodes} is the use of the augmented data matrices, obtained by transforming $\mathbf{X}$ and $\mathbf{X}'$ 
in $\widehat{\mathbf{X}}$ and  $\widehat{\mathbf{X}}'$, respectively:
\begin{equation}\label{eq:sXX'}
\widehat{\mathbf{X}}=
\begin{bmatrix}
\mathbf{X} \\ 
\mathbf{S}\\
\end{bmatrix},
\qquad
\widehat{\mathbf{X}}'=
\begin{bmatrix}
\mathbf{X}' \\ 
\mathbf{S}'\\
\end{bmatrix}.
\end{equation}
where the Hankel matrices $\mathbf{S}$, and $\mathbf{S}'$ are:  
\begin{equation}\label{eq:ss'}
\begin{split}
\mathbf{S}&=
\begin{bmatrix}
\mathbf{x}_{j-1} & \mathbf{x}_{j} & \dots & \mathbf{x}_{m-2}\\
\mathbf{x}_{j-2} & \mathbf{x}_{j-1} & \dots & \mathbf{x}_{m-3}\\
\vdots & \vdots & \vdots & \vdots \\
\mathbf{x}_{j-s-1} & \mathbf{x}_{j-s} & \dots & \mathbf{x}_{m-s-1}\\
\end{bmatrix}, \quad
\mathbf{S}'=
\begin{bmatrix}
\mathbf{x}_{j} & \mathbf{x}_{j+1} & \dots & \mathbf{x}_{m-1}\\
\mathbf{x}_{j-1} & \mathbf{x}_{j} & \dots & \mathbf{x}_{m-2}\\
\vdots & \vdots & \vdots & \vdots \\
\mathbf{x}_{j-s} & \mathbf{x}_{j-s+1} & \dots & \mathbf{x}_{m-s}\\
\end{bmatrix}.
\end{split}
\end{equation}

The augmentation of the system state with its delayed copies can be similarly applied to DMDc.
In this case, not only the state but also the input vector is extended with time-shifted copies, leading to the following representation of the dynamic system:
\begin{equation}\label{eq:hdmdcdsys}
    \mathbf{\hat{x}}_{j+1} = \mathrm{\mathbf{A}}\mathbf{\hat{x}}_j + \mathbf{B}\mathbf{\hat{u}}_j
\end{equation}
where $\hat{\mathbf{x}}_j$ follows the definition given for the Hankel-DMD, and $\hat{\mathbf{u}}_j = [\mathbf{u}_j,\,\mathbf{u}_{j-1},\,\dots\,,\mathbf{u}_{j-z}]^T$
is the extended input vector, including the $z$ delayed copies.
The formulation of the Hankel-DMDc can be obtained from the DMDc presented in \cref{s:dmdc} by transforming the matrices and $\mathbf{X}'$ in $\widehat{\mathbf{X}}'$ following \cref{eq:sXX',eq:ss'}, and $\mathbf{Y}$
in $\widehat{\mathbf{Y}}$ which is defined as: 
\begin{equation}\label{eq:scXX'}
\widehat{\mathbf{Y}}=
\begin{bmatrix}
\mathbf{X} \\
\mathbf{S}\\
\mathbf{U} \\
\mathbf{Z}\\
\end{bmatrix}
\end{equation}
and the matrix $\mathbf{Z}$ has a Hankel structure:  
\begin{equation}\label{eq:sts'}
\begin{split}
\mathbf{Z}&=
\begin{bmatrix}
\mathbf{u}_{j-1} & \mathbf{u}_{j} & \dots & \mathbf{u}_{m-2}\\
\mathbf{u}_{j-2} & \mathbf{u}_{j-1} & \dots & \mathbf{u}_{m-3}\\
\vdots & \vdots & \vdots & \vdots \\
\mathbf{u}_{j-z-1} & \mathbf{u}_{j-z} & \dots & \mathbf{u}_{m-z-1}\\
\end{bmatrix}.    
\end{split}
\end{equation}

The use of time-delayed copies as additional observables in the DMD has been connected to the Koopman operator as a universal linearizing basis \cite{Brunton2017}, yielding the true Koopman eigenfunctions and eigenvalues in the limit of infinite-time observations \cite{mezic2017}.

\subsection{Bayesian extension to Hankel-DMD and Hankel-DMDc}\label{s:bhdmd}
The shape and the values within matrices $\mathbf{A} \in \mathbb{R}^{ns \times ns}$ and $\mathbf{B} \in \mathbb{R}^{ns \times lz}$ show dependencies on three hyperparameters of the algorithms: the observation time length, $l_{tr} = t_m - t_1$ and the maximum delay time in the augmented state $l_{d_x} = t_{j-1} - t_{j-s-1}$ for Hankel-DMD, with the addition of the maximum delay time in the augmented input $l_{d_u} = t_{j-1} - t_{j-z-1}$ for Hankel-DMDc.

These dependencies can be denoted as follows:
\begin{equation}\label{eq:bayes1}
\begin{split}
&\mathrm{Hankel-DMD}: \quad\hspace{0.15cm} \mathbf{A}=\mathbf{A}(l_{tr},l_{d_x}) \\    
&\mathrm{Hankel-DMDc}: \quad \mathbf{A}=\mathbf{A}(l_{tr},l_{d_x},l_{d_u}), \qquad \mathbf{B}=\mathbf{B}(l_{tr},l_{d_x},l_{d_u}).
\end{split}
\end{equation}
In the Bayesian formulations, the hyperparameters are considered stochastic variables introducing uncertainty in the process.
Through uncertainty propagation, the solution $\mathbf{x}(t)$ also depends on $l_{tr}$, $l_{d_x}$ and, in Bayesian Hankel-DMDc, $l_{d_u}$.
At a given time $t$, the expected value of the solution and its standard deviation can be expressed for the Bayesian Hankel-DMD as:
\begin{equation}\label{eq:bayes3}
\boldsymbol{\mu_x}(t)=\int_{l_{d_u}^l}^{l_{d_u}^u} \int_{l_{d_x}^l}^{l_{d_x}^u} \mathbf{x}(t,l_{tr},l_{d_x})p(l_{tr})p(l_{d_x})dl_{tr} dl_{d_x},
\end{equation}
\begin{equation}\label{eq:bayes4}
\boldsymbol{\sigma_x}(t)= \left\{ \int_{l_{d_u}^l}^{l_{d_u}^u} \int_{l_{d_x}^l}^{l_{d_x}^u} 
  \left[\mathbf{x}(t,l_{tr},l_{d_x})-\boldsymbol{\mu_x}(t) \right]^2 p(l_{tr})p(l_{d_x})dl_{tr} dl_{d_x} \right\}^\frac{1}{2},
\end{equation}
and, for the Bayesian Hankel-DMDc as:
\begin{equation}\label{eq:bayes3c}
\boldsymbol{\mu_x}(t)=\int_{l_{d_u}^l}^{l_{d_u}^u} \int_{l_{d_x}^l}^{l_{d_x}^u} \int_{l_{tr}^l}^{l_{tr}^u}\mathbf{x}(t,l_{tr},l_{d_x},l_{d_u})p(l_{tr})p(l_{d_x})p(l_{d_u})dl_{tr} dl_{d_x} dl_{d_u},
\end{equation}
\begin{equation}\label{eq:bayes4c}
\boldsymbol{\sigma_x}(t)= \left\{ \int_{l_{d_u}^l}^{l_{d_u}^u} \int_{l_{d_x}^l}^{l_{d_x}^u} 
 \int_{l_{tr}^l}^{l_{tr}^u} \left[\mathbf{x}(t,l_{tr},l_{d_x})-\boldsymbol{\mu_x}(t) \right]^2 p(l_{tr})p(l_{d_x})p(l_{d_u})dl_{tr} dl_{d_x} dl_{d_u}\right\}^\frac{1}{2},
\end{equation}

where ${l_{tr}^l}$, ${l_{d_x}^l}$, ${l_{d_u}^l}$ and ${l_{tr}^u}$, ${l_{d_x}^u}$, ${l_{d_u}^u}$ are lower and upper bounds and $p(l_{tr})$, $p(l_{d_x})$, $p(l_{d_u})$, are the given probability density functions for $l_{tr}$, $l_{d_x}$ and $l_{d_u}$.

In practice, a uniform probability density function is assigned to the hyperparameters and a set of realizations is obtained through a Monte Carlo sampling. Accordingly, for each realization of the hyperparameters, the solution $\mathbf{x}(t,l_{tr},l_{d_x})$ or $\mathbf{x}(t,l_{tr},l_{d_x},l_{d_u})$ is computed, and at a given time $t$ the expected value and standard deviation of the solution are then evaluated.

The Bayesian extension is introduced to incorporate uncertainty quantification in the analyses, adding confidence intervals to the predictions coming from the numerical methods.

\section{Performance metrics}\label{s:metrics}
To evaluate the predictions made by the models and to compare the effectiveness of different configurations, three error indices are employed: the normalized mean square error (NRMSE) \cite{Diez2024}, the normalized average minimum/maximum absolute error (NAMMAE) \cite{Diez2024}, and the Jensen-Shannon divergence (JSD) \cite{marlantes2024}. 
All the metrics are averaged over the variables that constitute the system's state, providing a holistic assessment of prediction accuracy. This comprehensive evaluation considers aspects such as overall error, the range, and the statistical similarity of predicted versus measured values.

The NRMSE quantifies the average root mean square error between the predicted values $\mathrm{\mathbf{\tilde x}}_t$ and the measured (test) values $\mathrm{\mathbf{x}}_t$ at different time steps. 
It is calculated by taking the square root of the average squared differences, normalized by $k$ times the standard deviation of the measured values:
\begin{equation}\label{eq:nrmse}
   \mathrm{NRMSE} = \frac{1}{N} \sum_{i=1}^{N} \frac{1}{k\sigma_{x_i}}\sqrt{\frac{1}{\mathcal{T} } \sum_{j=1}^{\mathcal{T}} \left( \tilde{x}_{ij} - x_{ij} \right)^2},
\end{equation}
where $N$ is the number of variables in the predicted state, $\mathcal{T}$ is the number of considered time instants, and $\sigma_{x_i}$ is the standard deviation of the measured values in the considered time window for the variable $x_i$.

The NAMMAE metric, introduced in \cite{diez2022snh,Diez2024}, provides an engineering-oriented assessment of the prediction accuracy. It measures the absolute difference between the minimum and maximum values of the predicted and measured time series,  normalized by $k$ times the standard deviation of the measured values, as follows:
\begin{equation}\label{eq:nammae}
    \mathrm{NAMMAE} = \frac{1}{2 N } \sum_{i=1}^{N} \frac{1}{k\sigma_{x_i}}\left( \left| \min_j(\tilde{x}_{ij}) - \min_j(x_{ij}) \right| + \left| \max_j(\tilde{x}_{ij}) - \max_j(x_{ij}) \right| \right),
\end{equation}

Lastly, the JSD measures the similarity between the probability distributions of the predicted and reference signal \cite{marlantes2024}.
For each variable, it estimates the entropy of the predicted time series probability density function $Q$ relative to the probability density function of the measured time series $R$, where $M$ is the average of the two \cite{Lin1991}. 
\begin{align}
    &\mathrm{JSD} = \frac{1}{N} \sum_{i=1}^{N} \left( \frac{1}{2}D(Q_i\,||\,M_i) + \frac{1}{2}D(R_i\,||\,M_i) \right),  \quad \\
    &\quad \text{with} \quad M = \frac{1}{2} (Q + R)\label{eq:jsd}, \\
    &\quad \text{and} \quad D(K\,||\,H)=\sum_{y \in \chi} K(y) \ln\left( \frac{K(y)}{H(y)} \right). \label{eq:kld}
\end{align}
The Jensen-Shannon divergence is based on the Kullback-Leibler divergence $D$, given by \cref{eq:kld}, which is the expectation of the logarithmic difference between the probabilities $K$ and $H$, both defined over the domain $\chi$, where the expectation is taken using the probabilities $K$ \cite{Kullback1951}
The similarity between the distributions is higher when the Jensen-Shannon distance is closer to zero. JSD is upper bounded by $\ln(2)$.

Each of the three indices contributes to the error assessment with its peculiarity, providing a holistic assessment of prediction accuracy.
\begin{itemize}
    \item[-] The NRMSE evidences phase, frequency, and amplitude errors between the reference and the predicted signal, evaluating a pointwise difference between the two. However, it is not possible to discern between the three types of error and to what extent each type contributes to the overall value.
    \item[-] The NAMMAE indicates if the prediction varies in the same range of the original signal, but does not give any hint about the phase or frequency similarity of the two.
    \item[-] JSD index is ineffective in detecting phase errors between the predicted and the reference signals and is scarcely able to detect infrequent but large amplitude errors. Instead, it highlights whether the compared time histories assume each value in their range of variation a similar number of times. It is, hence, sensible to errors in the frequency and trend of the predicted signal.
\end{itemize}

An example of synergic use of the three is given for the case of a prediction that rapidly goes to zero and evolves with an overly small amplitude.
In this case, the NRMSE has a subtle behavior that may mislead the interpretation of the results if used alone: using the definition in \cref{eq:nrmse} the NRMSE would be close to 
an eighth of 
the standard deviation of the observed signal. This may be lower than, or comparable to, the error obtained with a prediction well capturing the trend of the observed time history but with a small phase shift, and may be misleading on the real capability of the algorithm at hand. 
The assessment of NAMMAE and JSD would help to discriminate the mentioned situation, as those metrics tackle different aspects of the prediction.

An additional time-resolved error index is considered, evaluating the time evolution of the squared root difference between the reference and predicted signal averaged among the variables and normalized by $k$ times the standard deviation of the measured values:
\begin{equation}\label{eq:varepsilon}
    \varepsilon(t) = \frac{1}{N} \sum_{i=1}^N \frac{1}{k\sigma_{x_i}}\sqrt{\left(\tilde{x_i}(t)-x_i(t)\right)^2}
\end{equation}
The index $\varepsilon(t)$ is used to investigate the progression of the prediction error in the test time frame and identify possible trends.

The value of $k$ in \cref{eq:nrmse,eq:nammae,eq:varepsilon} is set to 10, indicating a normalizing interval of $\pm 5 \sigma$ that corresponds to a coverage percentage equal to 96\% using Chebishev's inequality. 

\section{Numerical setup}\label{s:numset}
This section presents the numerical setup of the DMD for the nowcasting and system identification tasks, along with the preprocessing applied to the data. 
In this work, the analyses are performed on experimentally measured data, but it is worth noting the methods directly apply also to other data sources such as simulations of various fidelity levels.

A lexicon borrowed from machine learning can be used to describe the workflow for the DMD analyses due to its data-driven nature, even though the peculiarities of the method will cause some differences in the meaning of some terms.
Calling \textit{present instant} the DMD prediction starting point, the observed data lie in the past. 
The DMD models, hence, the matrices $\mathrm{\mathbf{A}}$ and $\mathrm{\mathbf{B}}$, are built using such past time histories that constitute the \textit{training data}. 
The \textit{test data}, conversely, lie in the future, and test sequences are used to assess the predictive performances of the models.

All DMD analyses are based on normalized data, using the Z-score standardization. Specifically, time histories are shifted and scaled with the average and variance evaluated on the training set.

\begin{figure}[ht!]   
    \centering
 \includegraphics[width=0.9\textwidth]{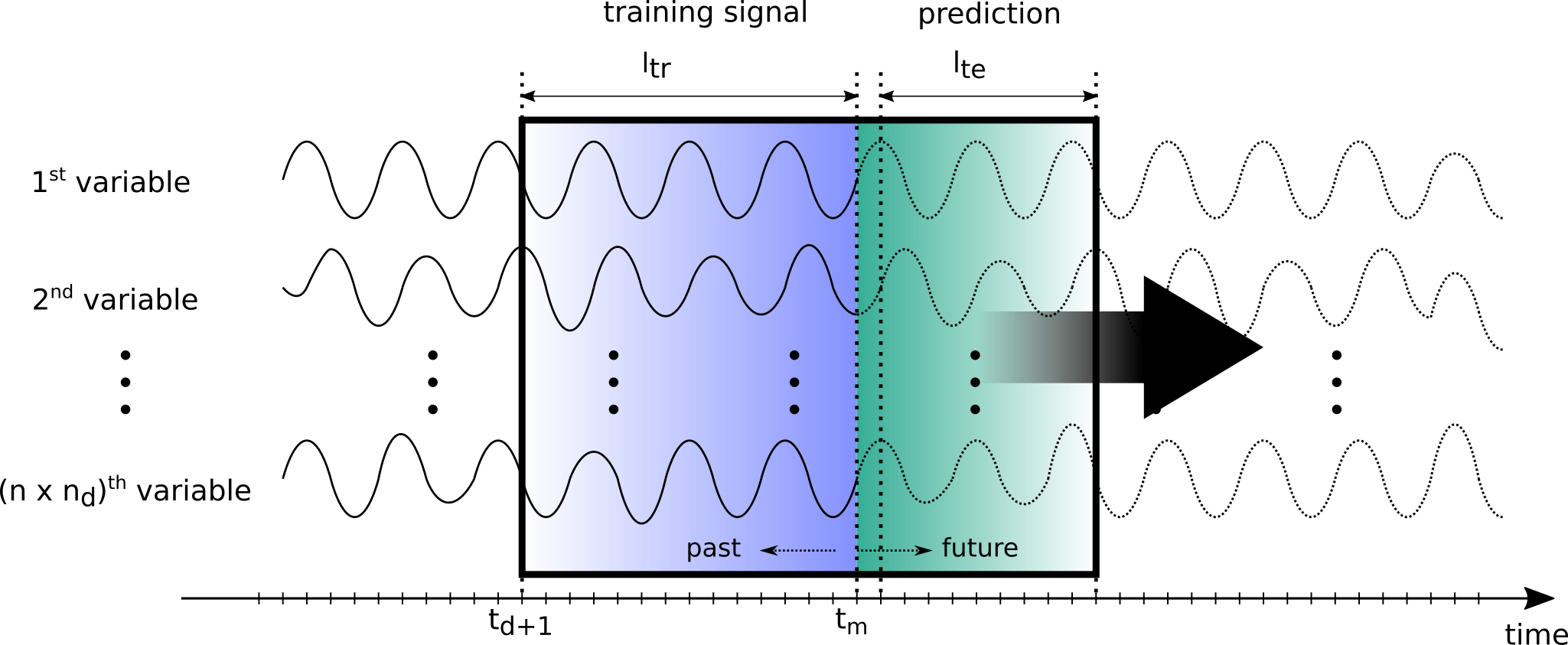}
    \caption{Sketch of the Hankel-DMD modeling approach for short-term forecasting (nowcasting.}
    \label{fig:sketchnowcasting}
\end{figure}

In the nowcasting approach, the (Bayesian) Hankel-DMD models are trained with sequences from the near-past, \textit{i.e.}, ending just before the present instant, and used to forecast short sequences in the future, in the order of few wave encounter periods (referring to the time between the passage of two consecutive waves relative to a fixed point on the floating platform). A new training is performed for each test sequence in a sliding window fashion, as sketched in \cref{fig:sketchnowcasting}.

\begin{figure}[ht!]
    \centering
 \includegraphics[width=\textwidth]{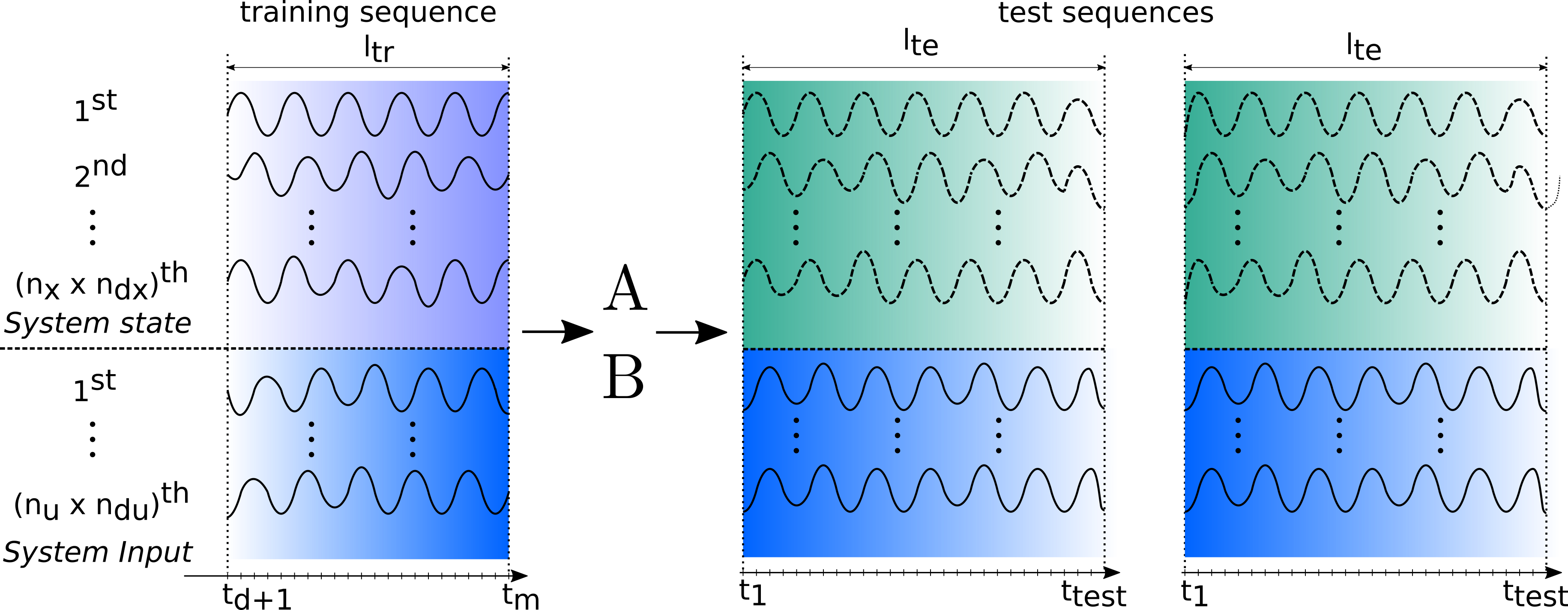}
    \caption{Sketch of the Hankel-DMD modeling approach for system identification.}
    \label{fig:sketchsi}
\end{figure}

In the system identification task, the training and the test phases are independent, such that a (Bayesian) Hankel-DMDc model is trained on a dedicated sequence only once and taken as representative ROM for the FOWT. Its performances are, hence, tested against multiple test sequences with no changes in the $\mathrm{\mathbf{A}}$ and $\mathrm{\mathbf{B}}$ matrices, as suggested by the sketch in \cref{fig:sketchsi}.

It is worth stressing that, differently from most machine learning methods, training a DMD/DMDc model is not an iterative procedure. In fact, the model is built, \textit{i.e.}, trained, with a direct procedure as described in \cref{s:dmd,s:dmdc}, identifying the DMD modes, the matrix $\mathrm{\mathbf{A}}$ and, for DMDc, the matrix $\mathrm{\mathbf{B}}$.

\begin{figure}[ht!]
\captionsetup[subfigure]{justification=centering}
  \begin{subfigure}[b]{0.45\linewidth}   
    \centering
    \includegraphics[width=\linewidth]{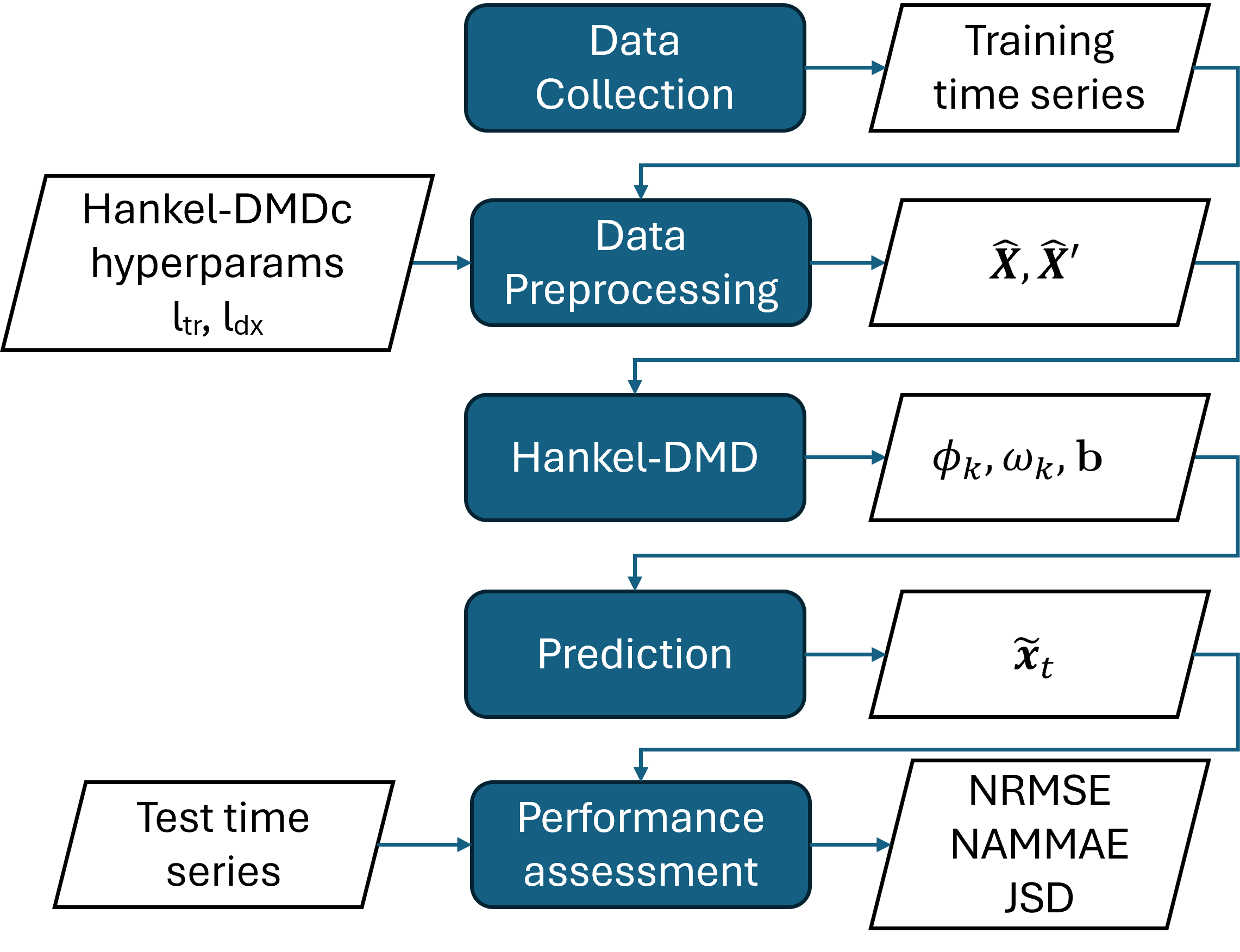}
    \caption{}\label{fig:wf_dmd}    
  \end{subfigure}
  \hfill
  \begin{subfigure}[b]{0.45\linewidth}   
    \centering
    \includegraphics[width=\linewidth]{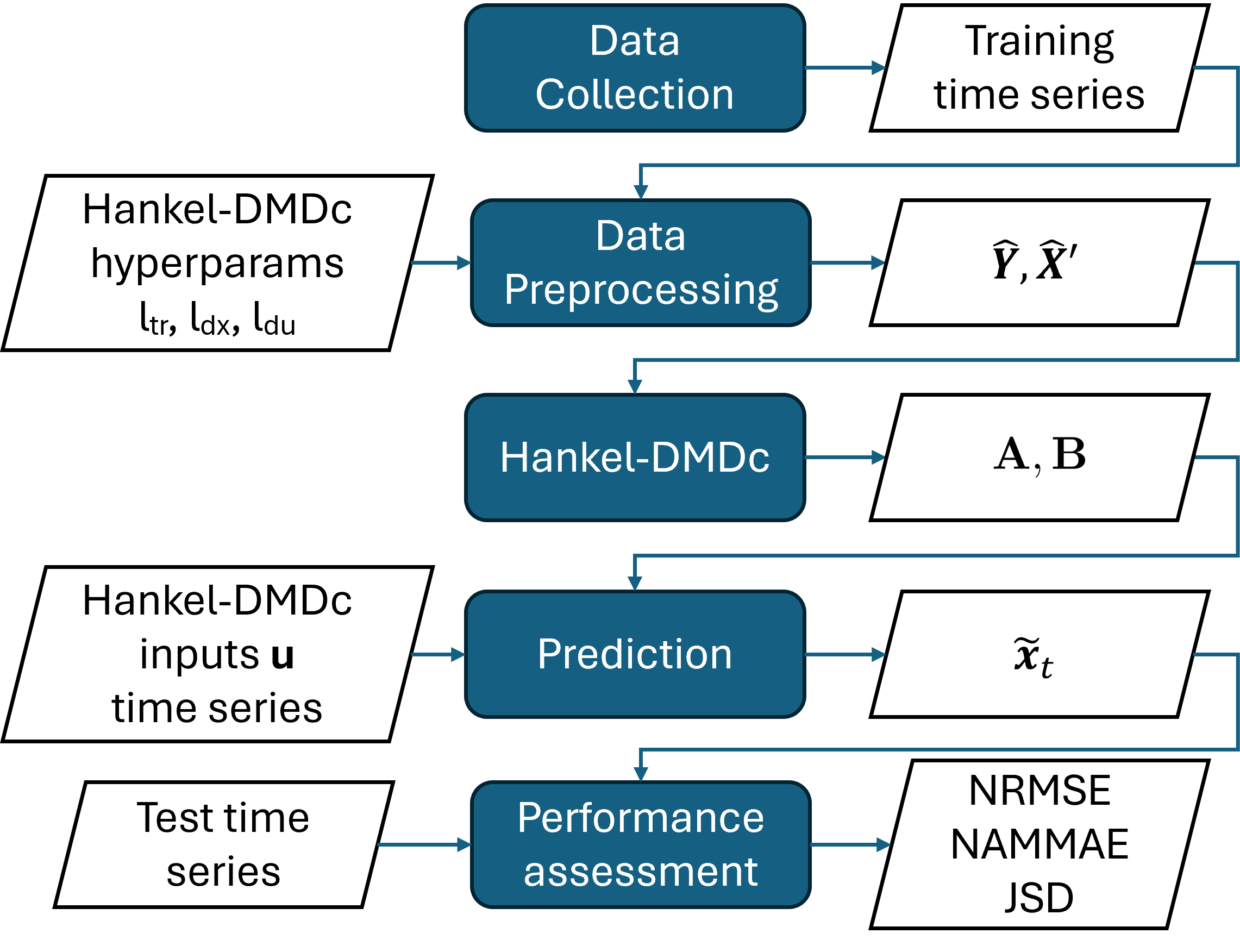}
    \caption{}\label{fig:wf_dmdc}    
  \end{subfigure}  
    \caption{Hankel-DMD (\subref{fig:wf_dmd}) and Hankel-DMDc (\subref{fig:wf_dmdc}) training-prediction-assessment flowchart}
    \label{fig:hdmdflowchart}
\end{figure}

\Cref{fig:hdmdflowchart} offers a view of the workflow of the Hankel-DMD/DMDc analyses. 
The first operation is the collection of the training data to be processed, which are in this work extracted from the existent dataset. 
Then, data are fed to the preprocessing step: time sequences are resampled using a sampling rate $dt=\hat{T}/32$ $s$ and then organized into the matrices $\mathrm{\mathbf{\hat X}}$ and $\mathrm{\mathbf{\hat X}}'$ based on hyperparameters values of the DMD method at hand.

The Hankel-DMD/DMDc algorithm is thus applied, obtaining the eigenvalues and eigenvectors of the $\mathrm{\mathbf{A}}$ matrix and the vector $\mathrm{\mathbf{b}}$ of the initial conditions for Hankel-DMD, or the matrices $\mathrm{\mathbf{A}}$ and $\mathrm{\mathbf{B}}$ for Hankel-DMDc.

The output of the Hankel-DMD or Hankel-DMDc is used to calculate the predicted time series $\mathbf{\tilde{x}}_t$ of the states, which are then compared with the test signals to evaluate the error metrics for performance assessment.

The nowcasting by Hankel-DMD only requires past histories of the variables under analysis, in contrast to the system identification in which the Hankel-DMDc needs the current value of the input time series at each time step to advance with the prediction. In particular, in this study, the wind speed $V_{w}$ and the wave elevation $h_{w}$ as measured by the PLC on the nacelle and the ACDP respectively are included in the vector $\mathbf{u}$.
The mentioned characteristics make the two approaches very different from each other. The nowcasting is more suitable for short-term predictions, in the range of few characteristic periods, and appropriate to be applied in real-time digital twins or model predictive controllers, also exploiting the possibility of continuous learning of the method during the system evolution.  
The system identification approach is more suited to produce a reduced order model of the system as a surrogate of the original one that produced the data used for training. 
Once trained, the ROM can be applied for fast and reliable predictions of the system's response to given operational conditions of possibly undefined time extension at a reduced computational cost, with potentially useful applications in control, design, life-cycle cost assessment, maintenance, and operational planning, as an alternative to high-fidelity simulations or experiments.

A full-factorial design-of-experiment is conducted to investigate the influence of the two main hyperparameters of the Hankel-DMD on the nowcasting performances. Five levels of variation are used for $l_{tr}={\hat T, 2\hat T, 4\hat T, 8\hat T, 16\hat T}$ and 6 for $l_{d_x}={0.5\hat T, \hat T, 2\hat T, 4\hat T, 8\hat T, 16 \hat T}$, as resumed in \cref{tab:doe} also in terms of training time steps $n_{tr}$ and number of delayed time histories embedded $n_d$ with the current time sampling.
\begin{table}[ht!]
    \centering
    \caption{List of the hyperparameter settings tested for Hankel-DMD nowcasting}\label{tab:doe}
    \begin{tabular}{llllllll}
     \toprule
      $l_{tr}$  & [-] &             & $\hat{T}$   & $2\hat{T}$ & $4\hat{T}$ & $8\hat{T}$ & $16\hat{T}$ \\
      $l_{d_x}$   & [-] & $0.5\hat T$ & $\hat{T}$   & $2\hat{T}$ & $4\hat{T}$ & $8\hat{T}$ & $16\hat{T}$ \\      
            \midrule
      $n_{tr}$  & [-] & 16          & 32 & 64  & 128 & 256 & 512 \\
      $n_{d_x}$ & [-] & 16          & 32 & 64  & 128 & 256 & 512 \\
      \bottomrule
    \end{tabular}
\end{table}
The prediction performance of each configuration is assessed through the NRMSE, NAMMAE, and JSD metrics introduced in \cref{s:metrics} on a statistical basis using 100 random starting instants as validation cases for prediction inside the 12-hour considered time frame. A forecasting horizon of $l_{te}=4\hat{T}$ is considered, corresponding to approximately 30s.

The statistical assessment of the system identification procedure shall also include the choice of the training sequence as an influencing parameter.
For this reason, the dataset is subdivided into training, validation, and test portions completely separated from each other. Ten different training sets are identified inside the data portion dedicated to training. For each one of them, the effect of the three hyperparameters of Hankel-DMDc on the system identification is analyzed with a full factorial design-of-experiment using seven levels for $l_{tr}$ and six levels for $l_{d_x}$ and $l_{d_u}$, using values in \cref{tab:doe_si}.
\begin{table}[ht!]
    \caption{List of the hyperparameter settings tested for Hankel-DMDc system identification algorithm}\label{tab:doe_si}
    \centering
    \begin{tabular}{lllllllll}
     \toprule
      $l_{tr}$ & [-] & $50\hat{T}$& $75\hat{T}$& $100\hat{T}$& $125\hat{T}$& $150\hat{T}$&$175\hat{T}$& $200\hat{T}$ \\
      $l_{d_x}$  & [-] & $0.5\hat{T}$ & $\hat{T}$   & $2\hat{T}$ & $3\hat{T}$ & $4\hat{T}$ & $5\hat{T}$ & \\  
      $l_{d_u}$  & [-] & $5\hat{T}$ & $10\hat{T}$   & $20\hat{T}$ & $30\hat{T}$ & $40\hat{T}$ & $50\hat{T}$ & \\        
            \midrule
      $n_{tr}$   & [-] & 1600 & 2400 & 3200 & 4000 & 4800 & 5600 & 6400\\
      $n_{d_x}$  & [-] & 16 & 32 & 64 & 96  & 128 & 160 &\\
      $n_{d_u}$  & [-] & 160 & 320 & 640 & 960  & 1280 & 1600 &\\
      \bottomrule
      \end{tabular}
\end{table}
The prediction performance of each of the 252 configurations is assessed through the NRMSE, NAMMAE, and JSD metrics on a statistical basis using 10 random validation signals extracted from the validation set, for a total of 100 evaluations of each error metric per hyperparameters configuration.
The prediction time window considered has a length of $l_{te}=20\hat{T}$, approximately 146s.

Results from the deterministic Hankel-DMD and Hankel-DMDc analyses are used to identify suitable ranges for the hyperparameter values to be used with their Bayesian extensions. 

\section{Results} \label{s:res}
This section discusses the DMD analysis of the system dynamics in terms of complex modal frequencies, modal participation, and most energetic modes, and assesses the results of the DMD-based forecasting method for the prediction of the state evolution.

\subsection{Modal analysis}
\begin{figure}[ht!]
\centering
\captionsetup[subfigure]{justification=centering}
    \begin{subfigure}[b]{0.46\linewidth}       
            \includegraphics[width=\hsize]{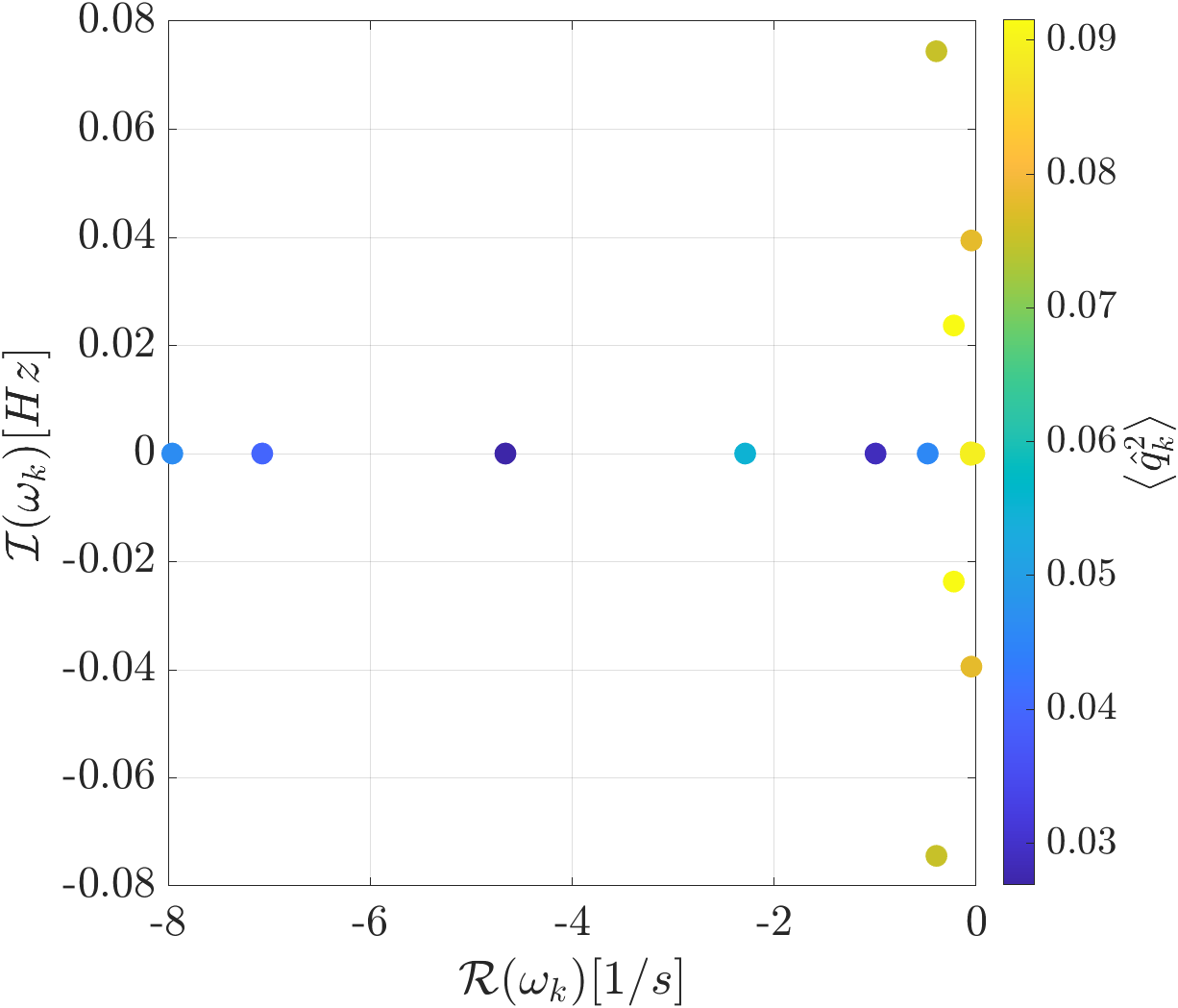}
            \caption{}\label{fig:dmd_a}
    \end{subfigure}       
    \hspace{1.0cm}
    \begin{subfigure}[b]{0.43\linewidth}       
            \includegraphics[width=\hsize]{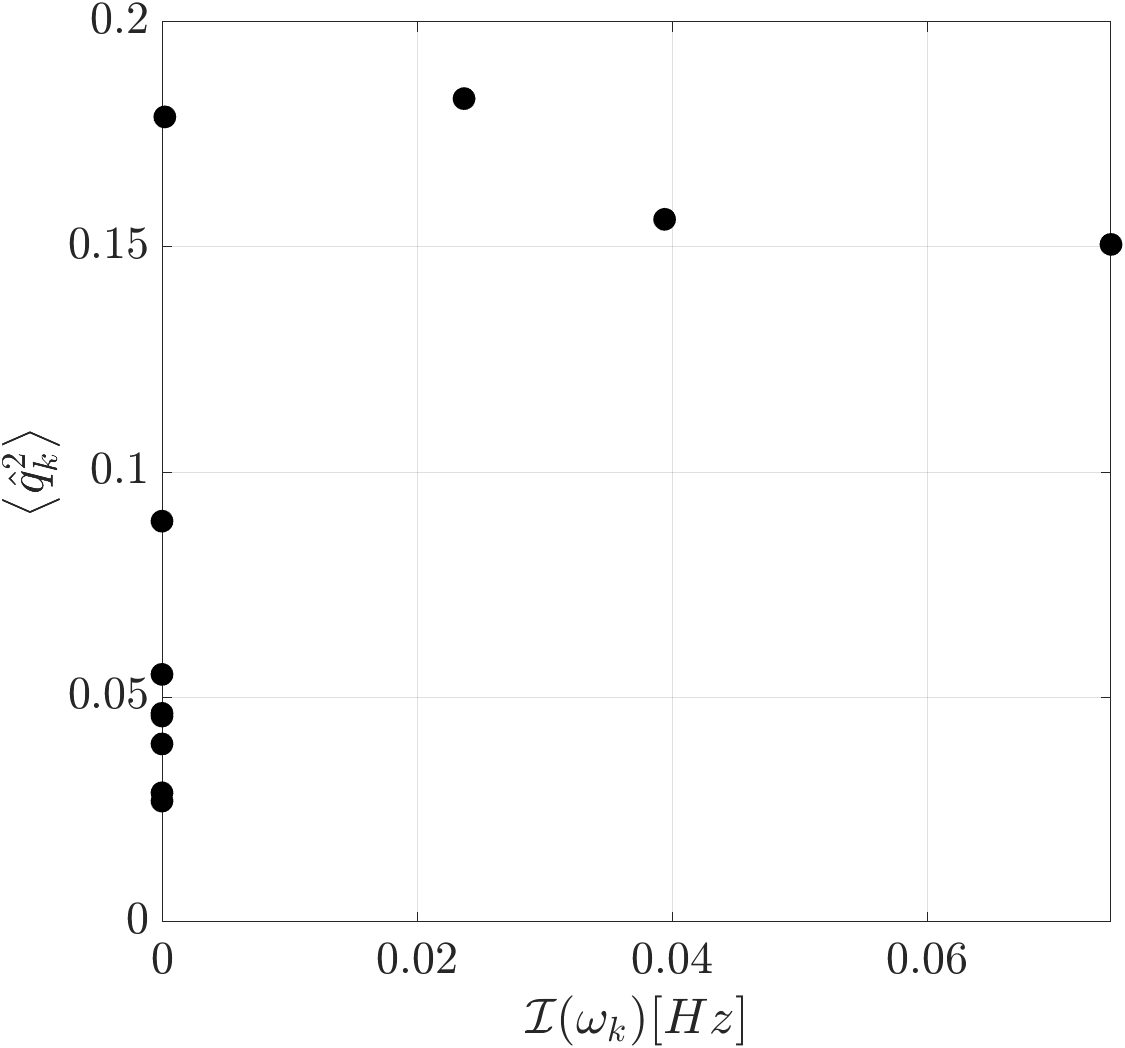}
            \caption{}\label{fig:dmd_b}
    \end{subfigure}    \\
    \vspace{0.05cm}
    \begin{subfigure}[b]{0.49\linewidth}
            \includegraphics[width=\hsize]{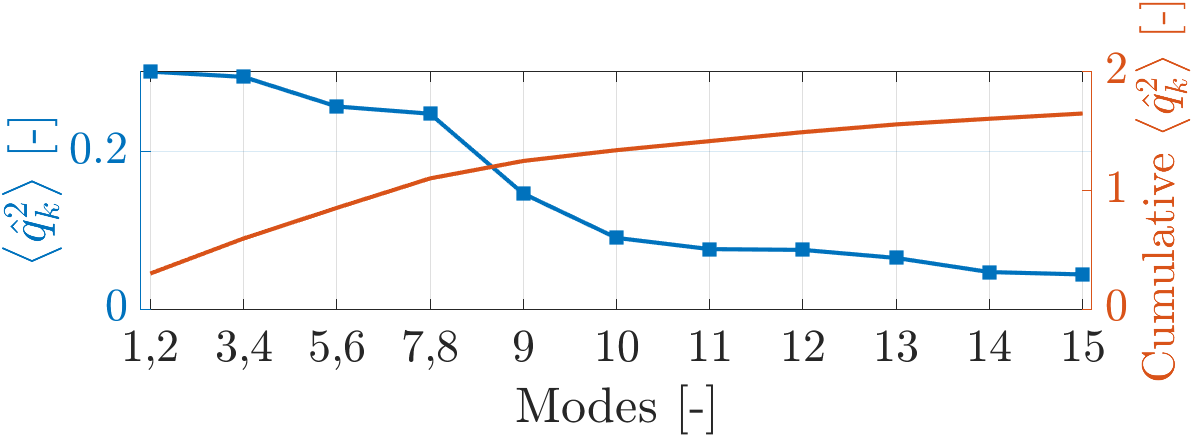}
            \caption{}\label{fig:dmd_c}
    \end{subfigure}
    \hfill
    \begin{subfigure}[b]{0.50\linewidth}            
            \includegraphics[width=\hsize]{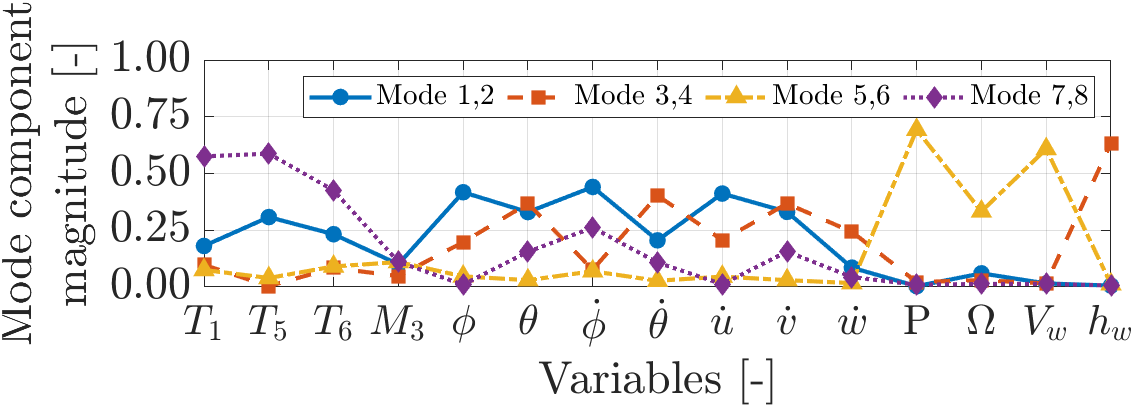}
            \caption{}\label{fig:dmd_d}    
    \end{subfigure}
\caption{DMD complex modal frequencies, modal participation, and first modes shapes.}\label{fig:dmd}    
\end{figure}

\Cref{fig:dmd} presents the DMD results for the FOWT system dynamics. 
Specifically, \cref{fig:dmd_a} shows the complex modal frequencies identified by DMD with no state augmentation.
These are ranked in \cref{fig:dmd_b} based on the normalized energy of the respective modal coordinate signal:
\begin{equation}
   \left\langle \hat{q}_k^2 \right\rangle = \frac{\left\langle {q}_k^2\right\rangle}{\sum_{k=1}^n\left\langle {q}_k^2 \right\rangle}, \quad\text{with}\quad \left\langle q_k^2 \right\rangle = \int_{t_i}^{t_f} q_k^2(t) dt.
\end{equation}
As can be seen in \cref{fig:dmd_c}, the dynamic is dominated by four couples of complex conjugate modes. Their components' magnitude are presented in \cref{fig:dmd_d}.
Modes (1,2) and (7,8) show a slow (frequency around 0.0237Hz, period around 42.3s) and a faster (frequency around 0.0745Hz, period around 13.5s) coupling between the loads on tendons (and weakly on moorings) and the platform motion. 
Modes (3,4) variable participation suggests the modes describe a slow wave height oscillation (frequency 0.00022Hz, with a period of almost 1 hour and 15 minutes) influencing the platform motion variables, both angular and linear.
Finally, modes (5,6) identify a subsystem describing the variation of extracted power and turbine rotational speed with wind speed (frequency 0.039Hz, period around 25.4s).
Interestingly, the power extracted by the wind turbine and the rpm of its blades are almost not involved in the description of the floating motions and seem scarcely influenced by them.
These results provide a solid background for choosing $h_w$ and $V_w$ as input variables for the system identification task.

\subsection{Nowcasting via Hankel-DMD}\label{s:nchdmd}
As anticipated in \cref{s:numset}, a full-factorial combination of settings of the Hankel-DMD hyperparameters is tested to investigate their influence on the forecasting capability of the method.

\begin{figure}[ht!]
\centering
\includegraphics[width=0.7\linewidth]{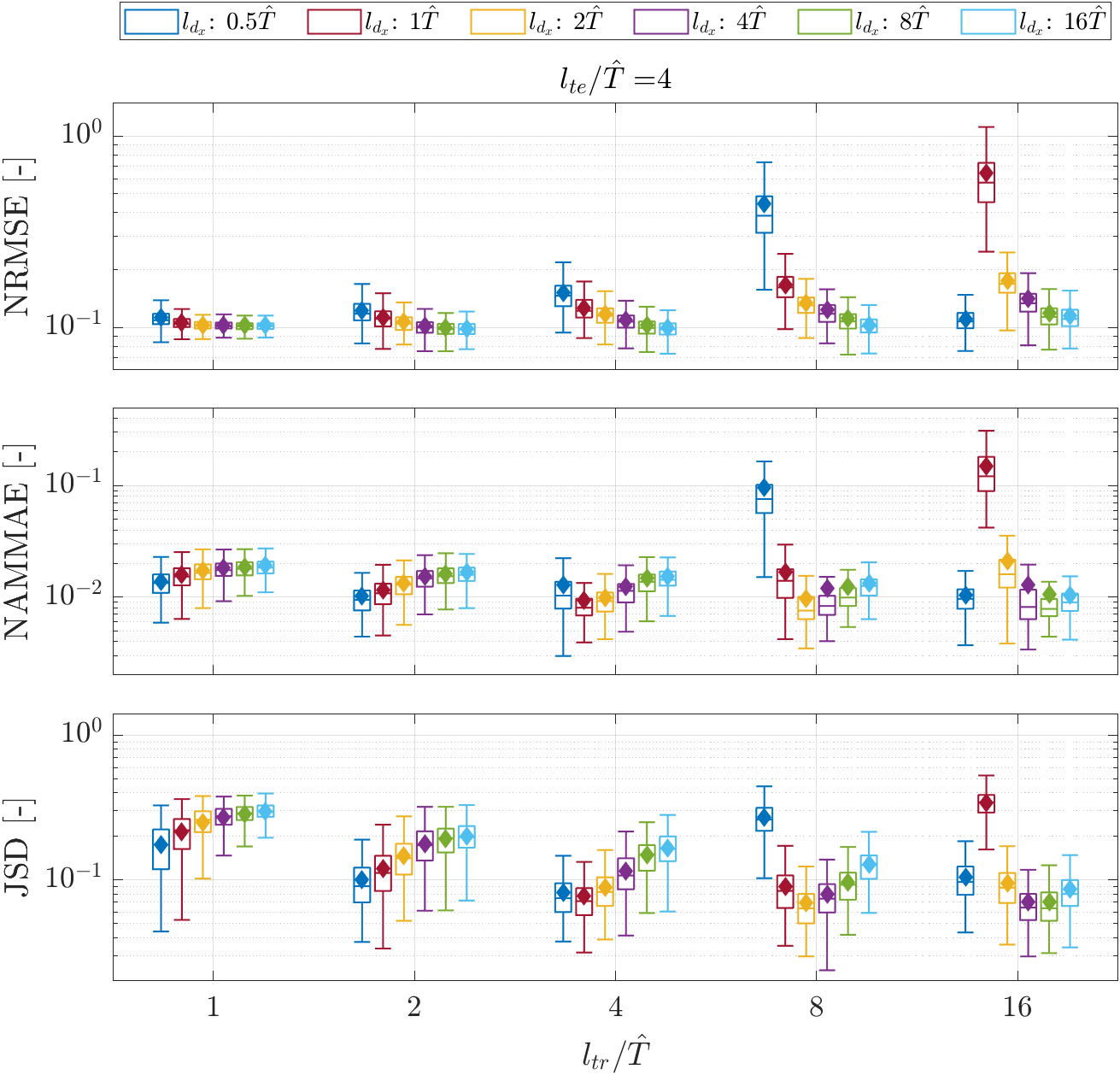}
\caption{Hankel-DMD, boxplot of error metrics over the validation set for tested $l_{tr}$ and $l_{d_x}$, $l_{te}=4\hat T$ ($\sim 30s$). Diamonds indicate the average value of the respective configuration.}\label{fig:ncmetrics} 
\end{figure}

The results, gathering the outcomes from all the algorithm setups and test cases, are resumed in \cref{fig:ncmetrics} as boxplots for the NRMSE, NAMMAE, and JSD, focusing on a length of the test signal $l_{te}=4\hat T$.
The boxplots show the first, second (equivalent to the median value), and third quartiles, while the whiskers extend from the box to the farthest data point lying within 1.5 times the inter-quartile range, defined as the difference between the third and the first quartiles from the box. 
The diamonds indicate the average of the results for each combination. 
Outliers are not shown to improve the readability of the plot.
It can be noted that the three metrics indicate different \textit{best} hyperparameter settings. 
Their values are reported in \cref{tab:detbestconfnc}.
However, the different types of errors targeted by the three metrics help to interpret the results and gain some insight:
\begin{enumerate}[a)]
    \item Long training signals with few delayed copies in the augmented state show poor prediction capabilities, as confirmed by all the metrics for both the short-term and mid-term time windows. The effect is notable for $\frac{l_{d_x}}{l_{tr}} < 1/8$. 
    \item A high number of embedded time-delayed signals with insufficiently long training lengths is prone to produce NRMSE progressively reducing its dispersion around the value of an eighth of the standard deviation of the observed signals. 
    This happens, with the explored values of $l_{d_x}$, particularly for $l_{tr}= \hat T$, $2\hat T$, and, to a less extent, $4 \hat T$, when $\frac{l_{d_x}}{l_{tr}}$ exceeds $1$. 
    At the same time, the NAMMAE and JSD values for the same settings are progressively increasing their values; this indicates that the predicted signals aren't able to catch the maximum and minimum values of the reference sequences and that the distribution of the predicted data is not adherent to the true data advancing in time.
    The combination of those two behaviors is due to the method generating numerous rapidly decaying predictions, which signals become null after a short observation time.
\end{enumerate}

A suitable range for the hyperparameter settings to obtain accurate results can be inferred by combining the above considerations. 
The best deterministic results are obtained when $4 \leq l_{tr}/{\hat T} \leq 16$ and $l_{d_x}/l_{tr}={1}/{4}$.
\begin{figure}
    \centering
    \includegraphics[width=\linewidth]{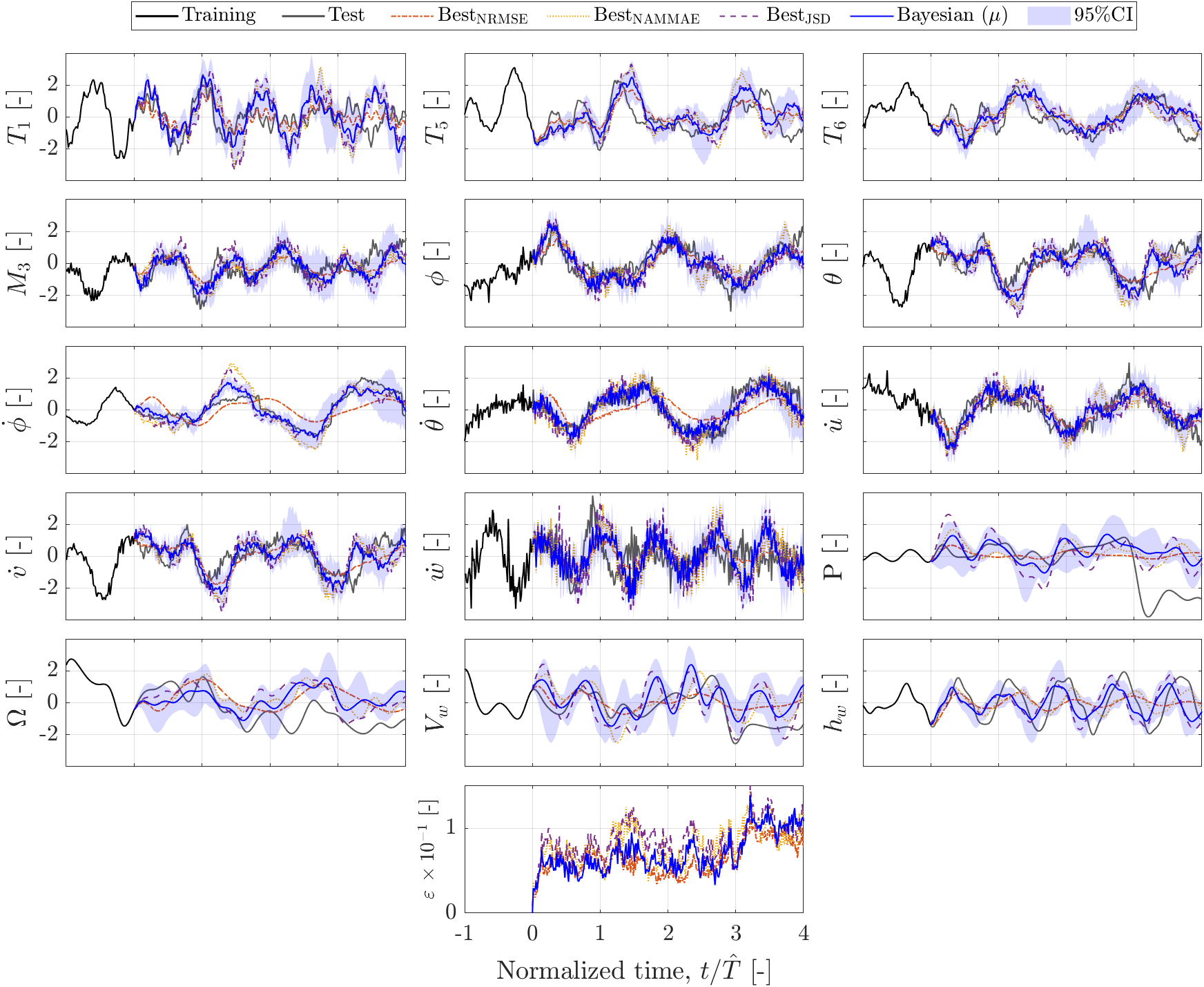}
    \caption{Standardized time series nowcasting by deterministic (hyperparameters for best average metrics) and Bayesian Hankel-DMD. Selected sequence 1.}
    \label{fig:pred66}
\end{figure}
\begin{figure}
    \centering
    \includegraphics[width=\linewidth]{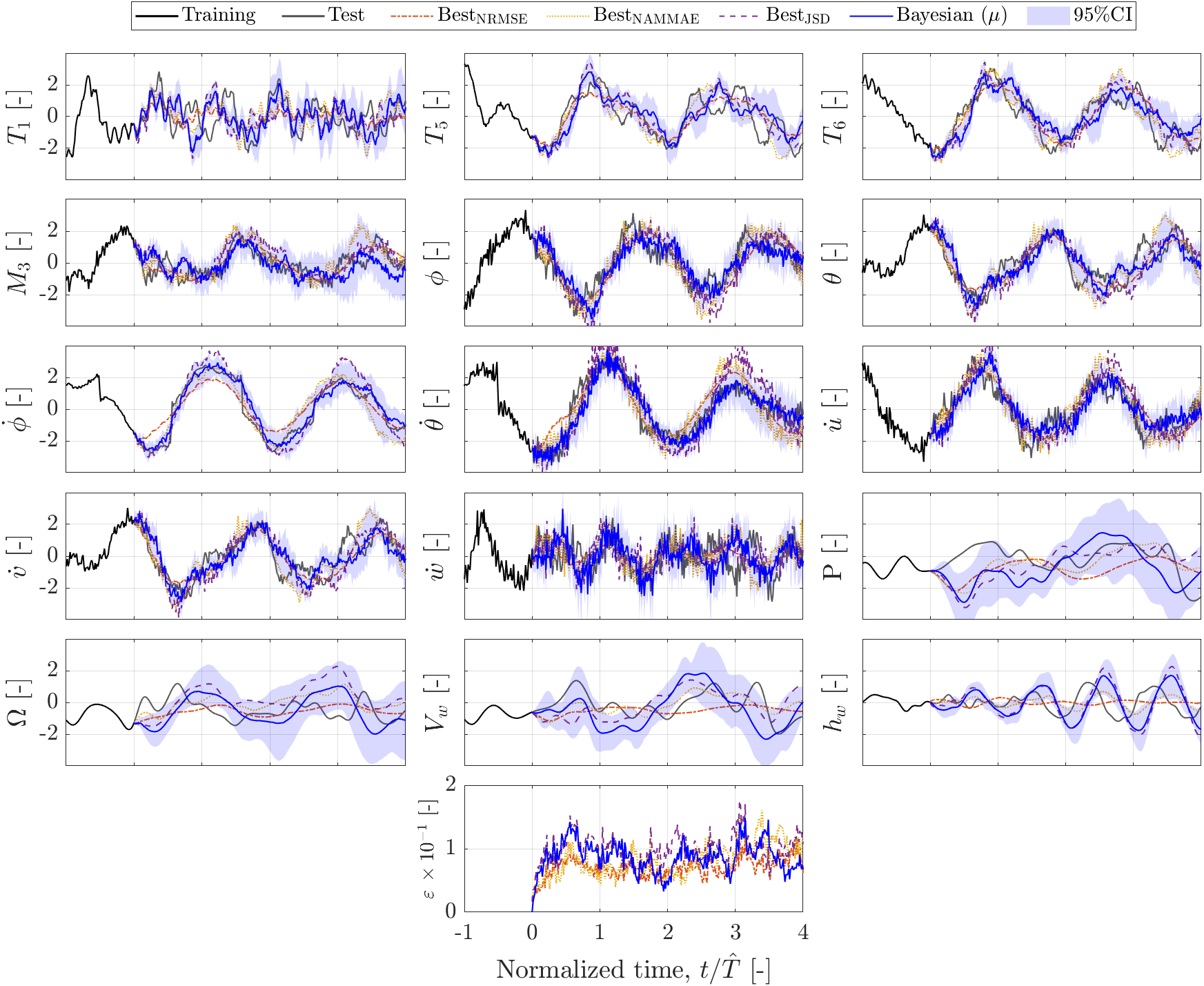}
    \caption{Standardized time series nowcasting by deterministic (hyperparameters for best average metrics) and Bayesian Hankel-DMD. Selected sequence 2.}
    \label{fig:pred128}
\end{figure}
\begin{figure}
    \centering   
    \includegraphics[width=\linewidth]{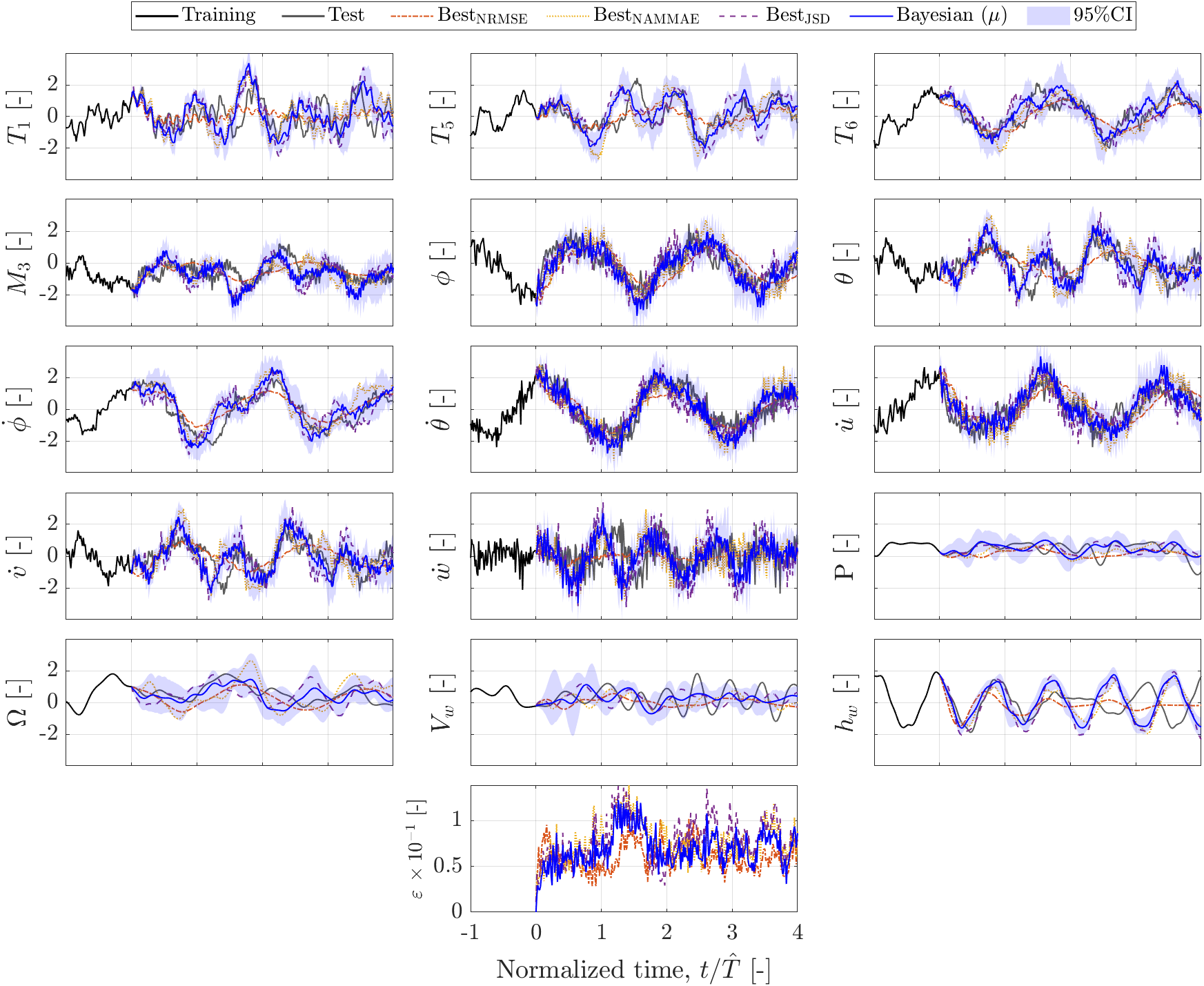}
    \caption{Standardized time series nowcasting by deterministic (hyperparameters for best average metrics) and Bayesian Hankel-DMD. Selected sequence 3.}
    \label{fig:pred185}
\end{figure}
\begin{figure}
    \centering
    \includegraphics[width=\linewidth]{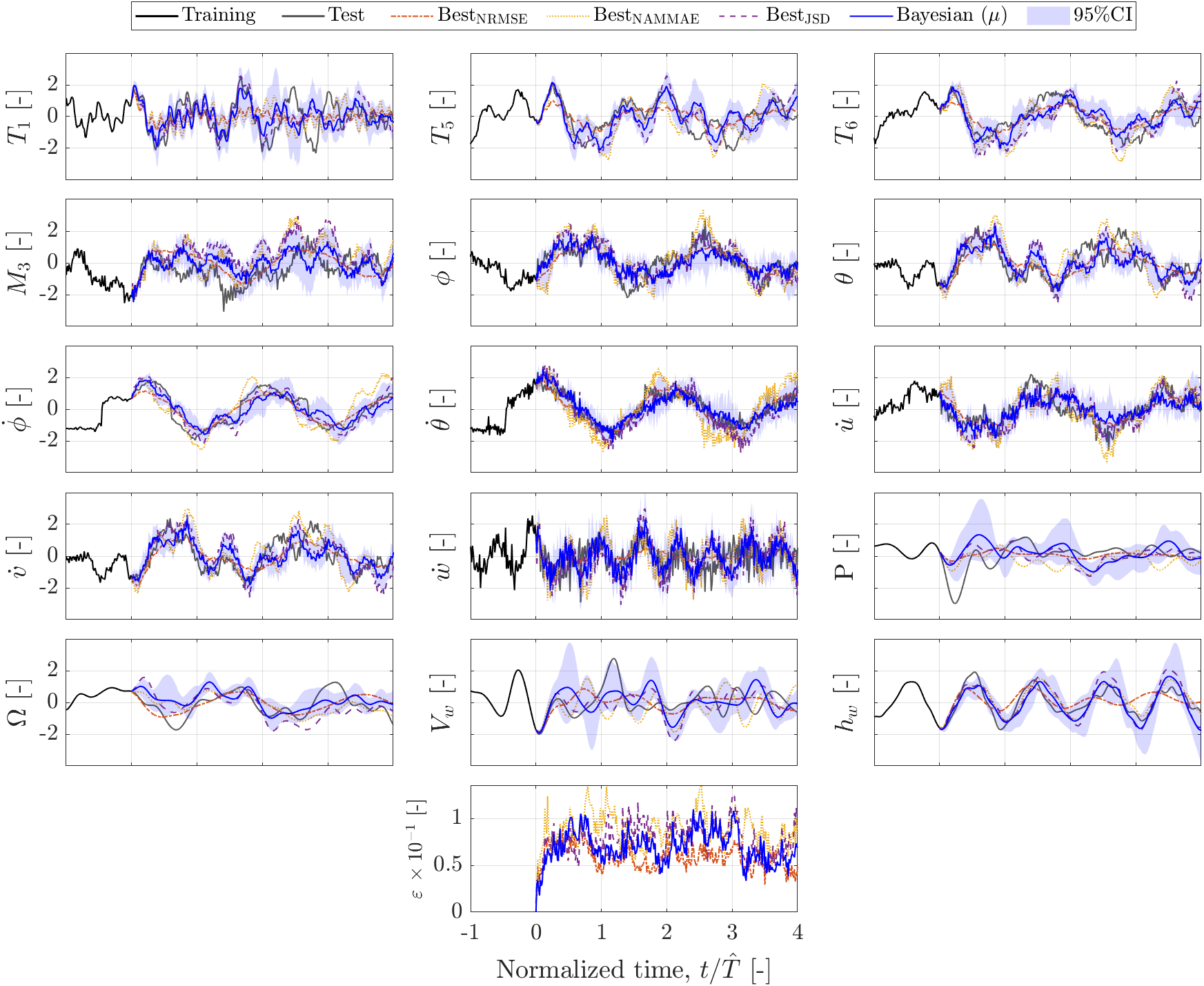}
    \caption{Standardized time series nowcasting by deterministic (hyperparameters for best average metrics) and Bayesian Hankel-DMD. Selected sequence 4.}
    \label{fig:pred240}
\end{figure}

\Cref{fig:pred66,fig:pred128,fig:pred185,fig:pred240} show the forecast by the Hankel-DMD for random test sequences taken as representative nowcasting examples. 
The figures show the last part of the training sequence with a solid black line, the test sequence in a dashed black line, and the prediction obtained with the Best$_\text{NRMSE}$, Best$_\text{NAMMAE}$ and Best$_\text{JSD}$ hyperparameters, as reported in \cref{tab:detbestconfnc}, with an orange dash-dotted, yellow dotted and purple dashed line respectively. 
Forecast data are in fairly good agreement with the measurements, particularly for the Best$_\text{NAMMAE}$ and Best$_\text{JSD}$ lines. 
Confirming the previous statement, some predictions of the Best$_\text{NRMSE}$ show a rapidly decaying amplitude, not following the ground truth, see \textit{e.g.,} $T_1$, $T_5$, $\theta$, $\dot{\phi}$, $\dot{v}$, $\dot{w}$, and $h_w$ in \cref{fig:pred185}, or $\dot{\phi}$, $\dot{\theta}$ in \cref{fig:pred66}.

As expected from the nowcasting algorithm, the forecasting accuracy is higher at the beginning of the prediction, but in most cases is satisfactory for the entire window.
The variables reproduced with the highest errors are the extracted power $P$, the blade rotational speed $\Omega$ and the wind speed $V_w$. 
Their time histories estimated by the PLC on the nacelle do not show strong periodicity in the observed time frame and, moreover, show strong non-linearities: the turbine control algorithms limit the maximum power extracted by the turbine and its blades' rotational speed, for example acting as a saturator when the wind is too strong.
In addition, and partially as a consequence of the above, these variables are seen to form an almost separate subsystem in the modal analysis; hence, a limited portion of the data can be actually used by the DMD to extract related modal content and produce their forecast, posing the method in a very challenging situation.

\subsection{Nowcasting via Bayesian Hankel-DMD}\label{s:ncbhdmd}
As highlighted by the authors in previous works \cite{diez2022snh, serani2023, Diez2024, serani2024snh}, the final prediction from DMD-based models may strongly vary for different hyperparameters settings, and no general rule is given for the determination of their optimal values. 
Aiming at including uncertainty quantification in the prediction, and making the prediction more robust, the Bayesian extension of the Hankel-DMD considers the hyperparameters of the method as stochastic variables with uniform probability density functions.

The insights gained in the deterministic analysis are applied to determine suitable variation ranges for $l_{tr}$ and $l_{d_x}$.
In particular, a probabilistic length of the training time history is considered, uniformly distributed between 4 to 16 incoming wave periods, $l_{tr}/{\hat T} \sim \mathcal{U}(4,16)$. 
Moreover, for each realization of $l_{tr}$, $l_{d_x}$ is also taken as a probabilistic variable, uniformly distributed in the interval 
$l_{d_x} \sim \mathcal{U}\left(\frac{l_{tr}}{8},l_{tr}\right)$ 
(each actual $n_d$ is taken as its respective integer part).

The solid blue lines in \cref{fig:pred66,fig:pred128,fig:pred185,fig:pred240} show the expected value of the Bayesian predictions on representative test set sequences, obtained using 100 uniformly distributed Monte Carlo realizations; the blue shadowed areas indicate the 95\% confidence interval of the uncertain predictions. 

The Bayesian Hankel-DMD provides fairly good forecasting of the measured quantities and the same considerations made in the deterministic analysis hold.
The biggest discrepancies are noted in the prediction of $P$, $\Omega$, and $V_w$, which also show the largest uncertainty bands, suggesting poor accuracy.

\begin{figure}[ht!]
\centering
\includegraphics[width=0.5\linewidth]{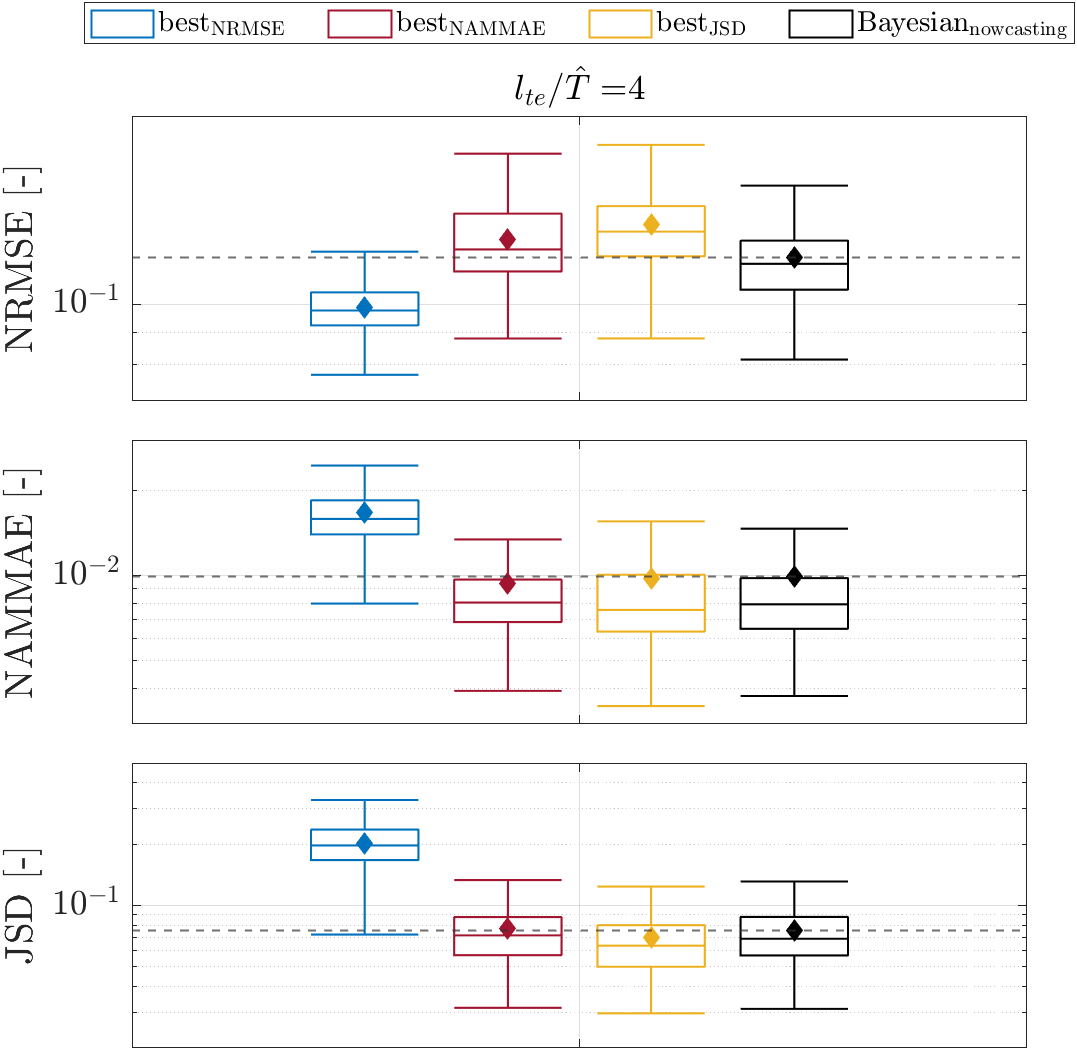}
\caption{Error metrics comparison, Hankel-DMD vs. Bayesian Hankel-DMD nowcasting, $l_{te}=4\hat T$ ($\sim 30s$).}\label{fig:ncdetvsbay}
\end{figure}

\Cref{fig:ncdetvsbay} compares the error metrics from the Bayesian Hankel-DMD and the Best$_\text{NRMSE}$, Best$_\text{NAMMAE}$ and Best$_\text{JSD}$ configurations of the deterministic analysis. 
for $l_{te}/\hat{T}=4$. The mean values of the metrics are summarized in \cref{tab:detbestconfnc}.
The boxplots indicate that Bayesian nowcasting achieves results comparable to the best deterministic configurations for NAMMAE and JSD, slightly improving their NRMSE. 
This appears remarkable considering the poor NAMMAE and JSD performance of the Best$_\text{NRMSE}$ configuration.
\begin{table}
\caption{Resume of the best hyperparameters configurations for nowcasting as identified by the deterministic design of experiment and Bayesian setup, $l_{te}=4 \hat{T}$.}\label{tab:detbestconfnc}
\begin{tabular}{llllll}
\toprule
                                   & NRMSE & NAMMAE & JSD   & $l_{tr}$       & $l_{d_x}$          \\
                                   & (avg) &  (avg) & (avg) &                &                               \\
      \midrule
       $\text{Best}_\text{NRMSE}$  &0.124 &0.026 &0.200 & 2$\hat{T}$     & 16$\hat{T}$         \\
       $\text{Best}_\text{NAMMAE}$ &0.159 &0.015 &0.077 & 4$\hat{T}$     & $\hat{T}$         \\
       $\text{Best}_\text{JSD}$    &0.168 &0.015 &0.070 & 8$\hat{T}$     & 2$\hat{T}$         \\       
       Bayesian                    &0.148 &0.015 &0.075 & [4-16]$\hat{T}$ & $l_{tr}/4$ \\
     \midrule
\end{tabular}
\end{table}

It shall be noted that the evaluation time for a single Hankel-DMD training, averaged over 10 realizations using 100 different hyperparameter values in the intervals of the Bayesian analysis, ranges from $\mu_t=0.029 s$, $\sigma_t=0.0041 s$ to $\mu_t=1.225 s$, $\sigma_t=0.0159 s$ on a mid-end laptop with an Intel Core i5-1235U CPU and 16Gb of memory using a non-compiled MATLAB code, depending on the algorithm setup (longer training signals with more delays require more time). 
The one-shot and computationally inexpensive training phase, related to direct linear algebra operations as detailed in \cref{s:dmd}, makes the Hankel-DMD and Bayesian Hankel-DMD algorithms very promising in the context of real-time forecasting and nowcasting. 
The forecasting time window here considered, \textit{i.e.,} $l_{te}=4\hat{T}$  ($\sim 30s$), can be assumed satisfactory from the perspective of the design of a model predictive or feedforward controller according to \cite{ma2018}, where a five-second horizon is considered and found to be sufficient. 
Moreover, the nowcasting algorithm is inherently suitable for continuous learning and digital twinning, adapting the model and predictions along with the incoming data from an evolving changing system and environment.

\subsection{System identification via Hankel-DMDc}\label{s:sihdmdc}
\begin{figure}
    \centering
\includegraphics[width=0.625\linewidth]{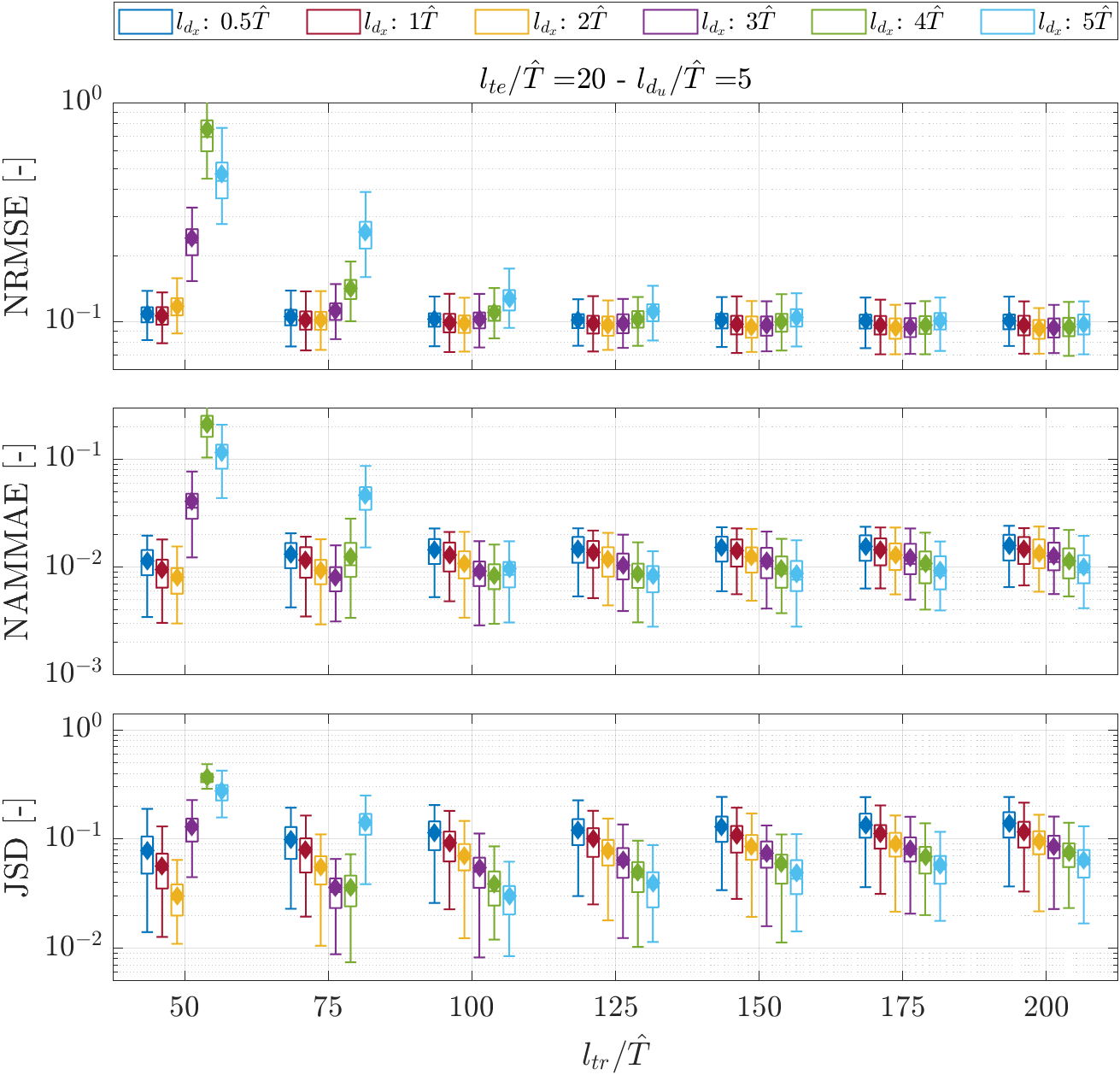}
    \caption{Hankel-DMDc, boxplot of error metrics over the validation set for tested $l_{tr}$ and $l_{d_x}$ with $l_{d_u}/\hat{T}=5$. Diamonds indicate the average value of the respective configuration. $l_{te}=20\hat{T}$ ($\sim 146s$).}
    \label{fig:simet1}
\end{figure}

\begin{figure}
    \centering    \includegraphics[width=0.625\linewidth]{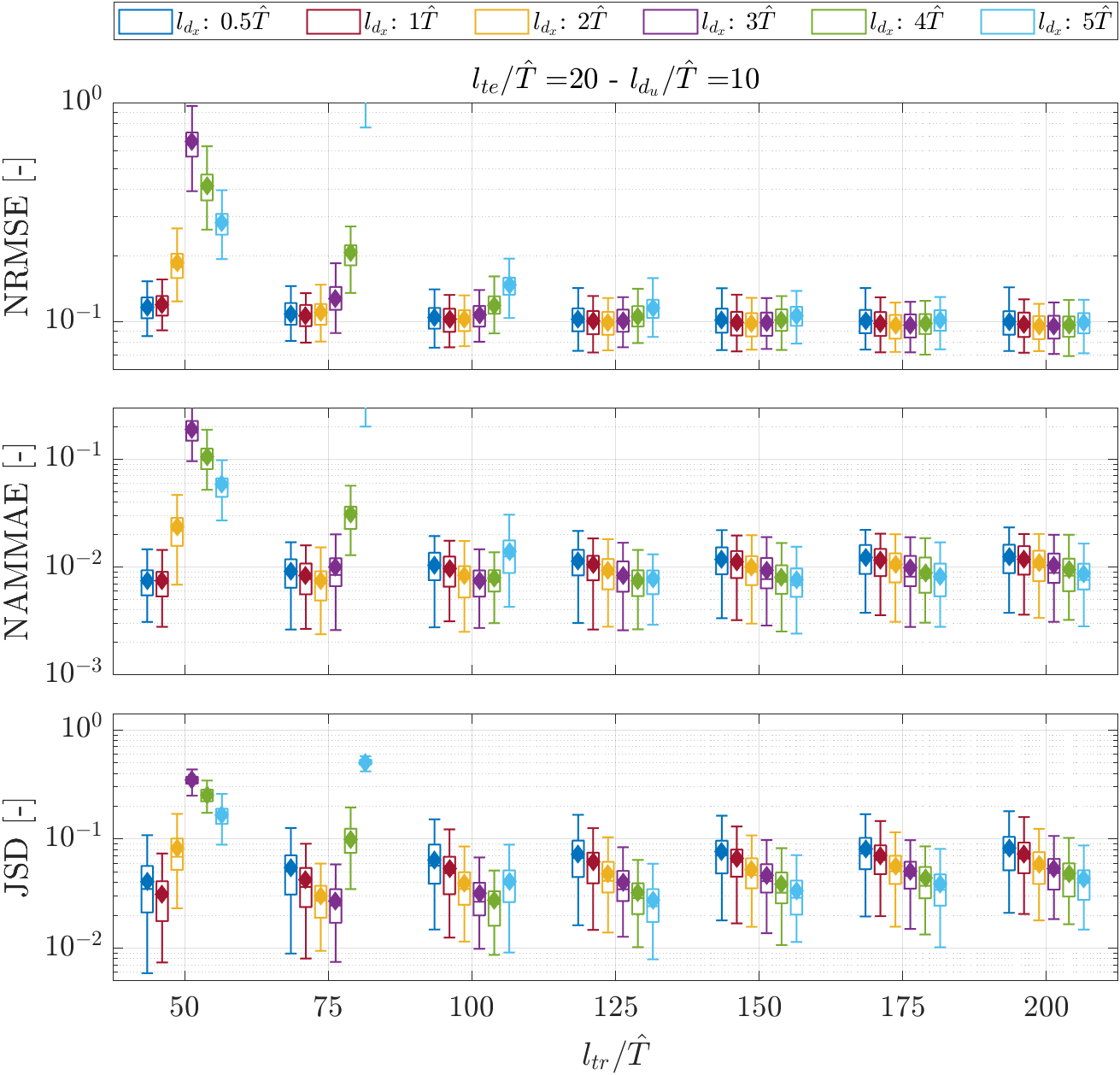}
    \caption{Hankel-DMDc, boxplot of error metrics over the validation set for tested $l_{tr}$ and $l_{d_x}$ with $l_{d_u}/\hat{T}=10$. Diamonds indicate the average value of the respective configuration. $l_{te}=20\hat{T}$ ($\sim 146s$).}
    \label{fig:simet2}
\end{figure}

\begin{figure}
    \centering
\includegraphics[width=0.625\linewidth]{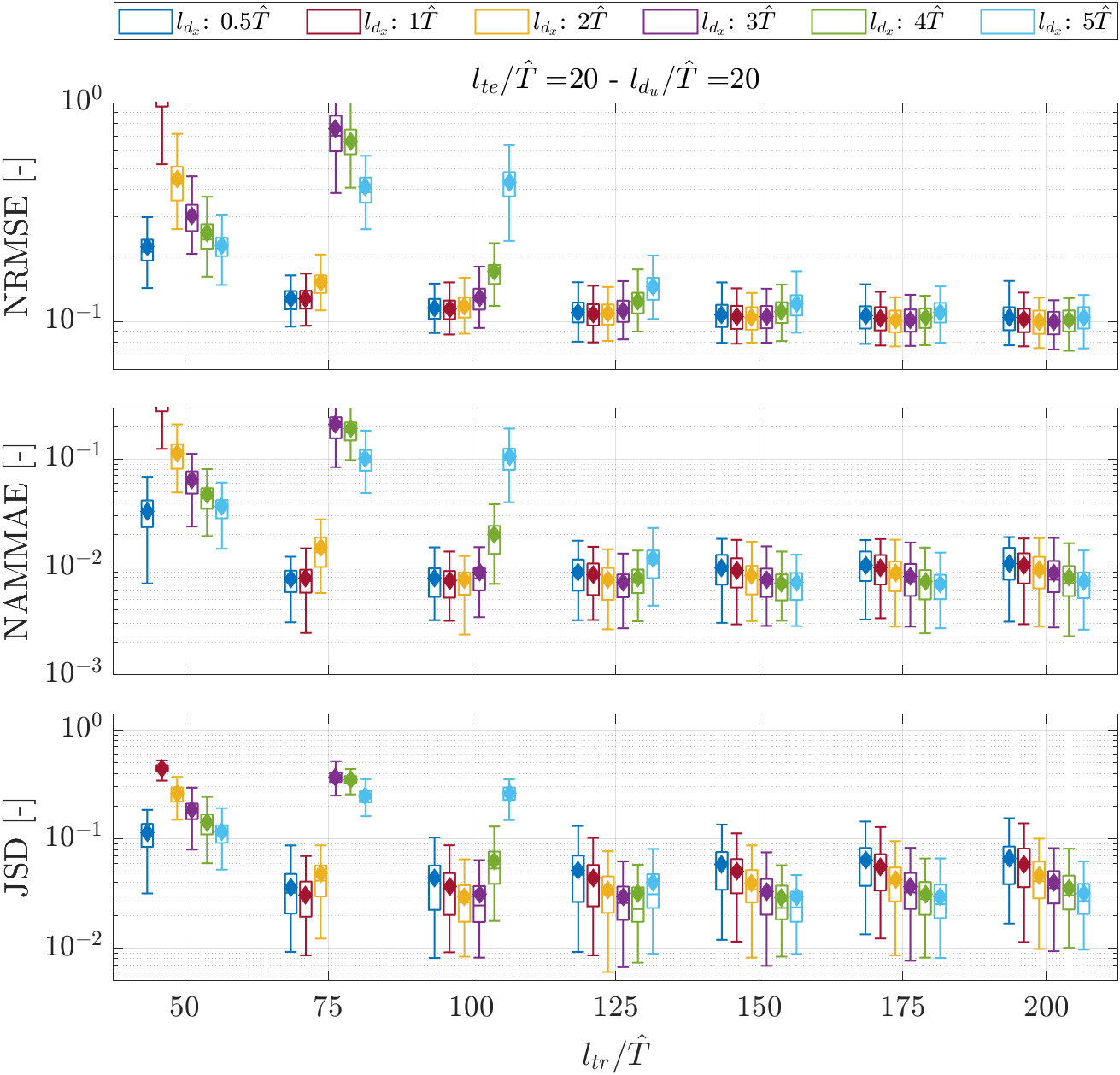}
    \caption{Hankel-DMDc, boxplot of error metrics over the validation set for tested $l_{tr}$ and $l_{d_x}$ with $l_{d_u}/\hat{T}=20$. Diamonds indicate the average value of the respective configuration. $l_{te}=20\hat{T}$ ($\sim 146s$).}
    \label{fig:simet3}
\end{figure}

\begin{figure}
    \centering    \includegraphics[width=0.625\linewidth]{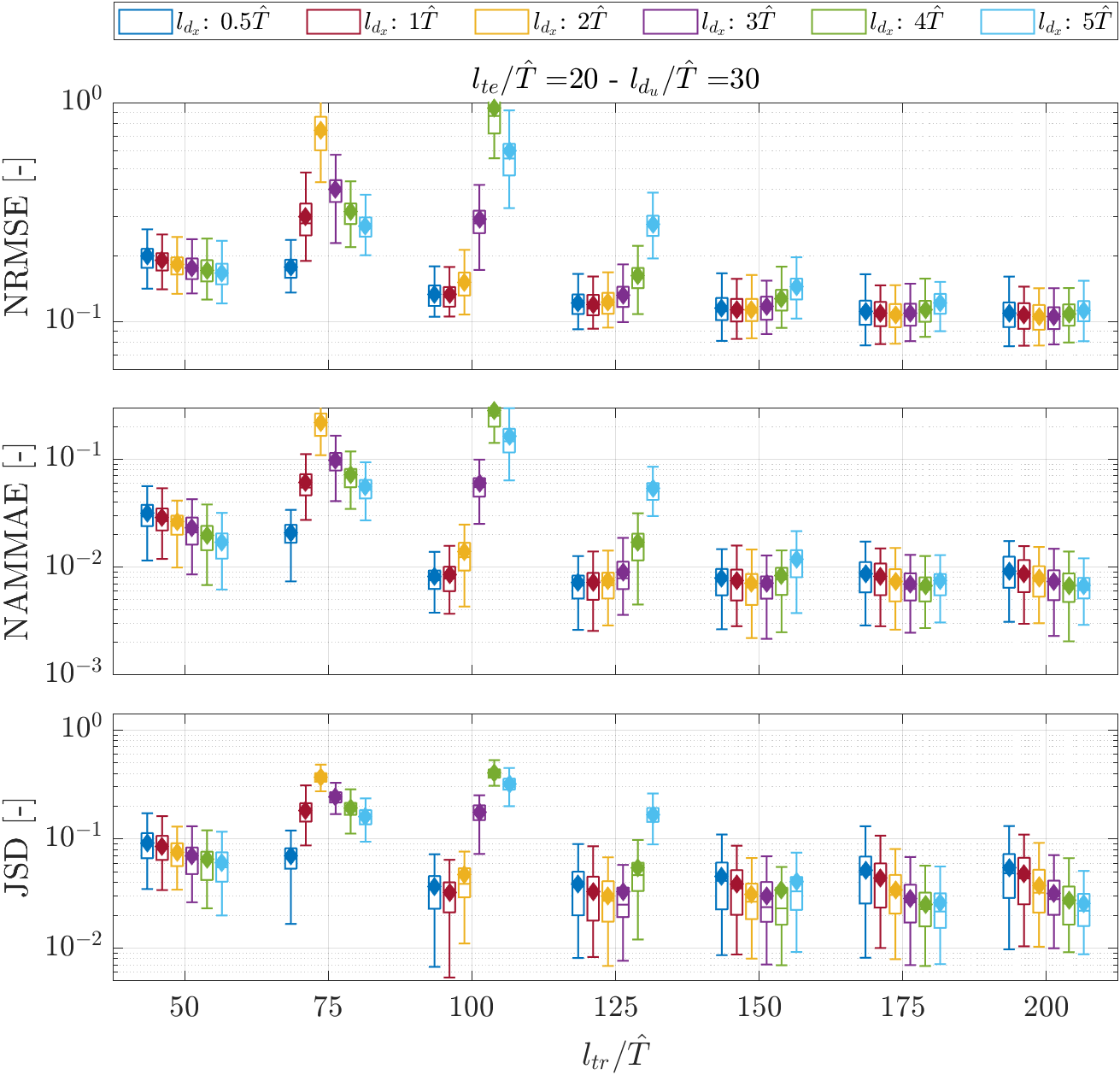}
    \caption{Hankel-DMDc, boxplot of error metrics over the validation set for tested $l_{tr}$ and $l_{d_x}$ with $l_{d_u}/\hat{T}=30$. Diamonds indicate the average value of the respective configuration. $l_{te}=20\hat{T}$ ($\sim 146s$).}
    \label{fig:simet4}
\end{figure}

\begin{figure}
    \centering
\includegraphics[width=0.625\linewidth]{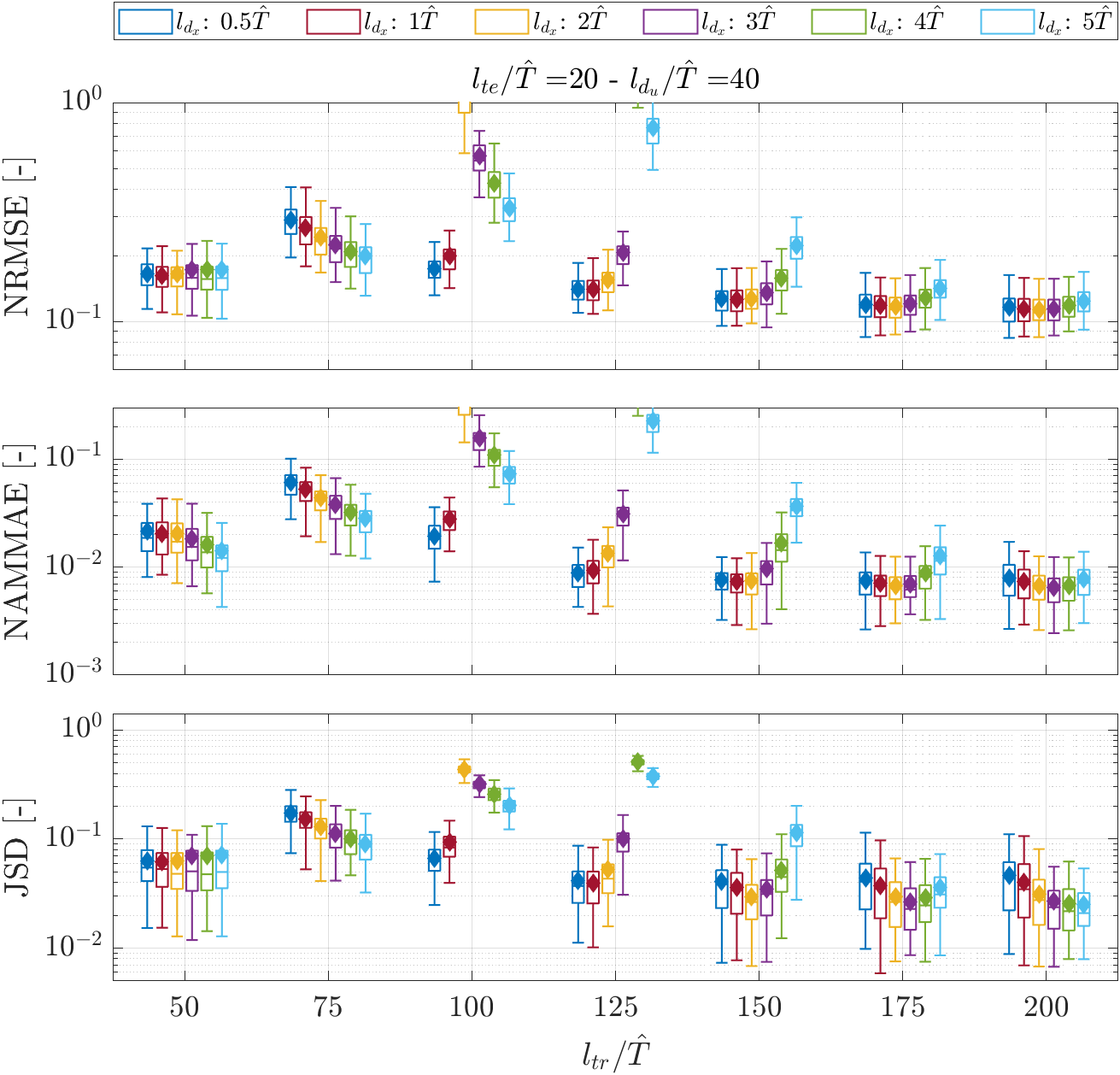}
    \caption{Hankel-DMDc, boxplot of error metrics over the validation set for tested $l_{tr}$ and $l_{d_x}$ with $l_{d_u}/\hat{T}=40$. Diamonds indicate the average value of the respective configuration. $l_{te}=20\hat{T}$ ($\sim 146s$).}
    \label{fig:simet5}
\end{figure}

\begin{figure}
    \centering    \includegraphics[width=0.625\linewidth]{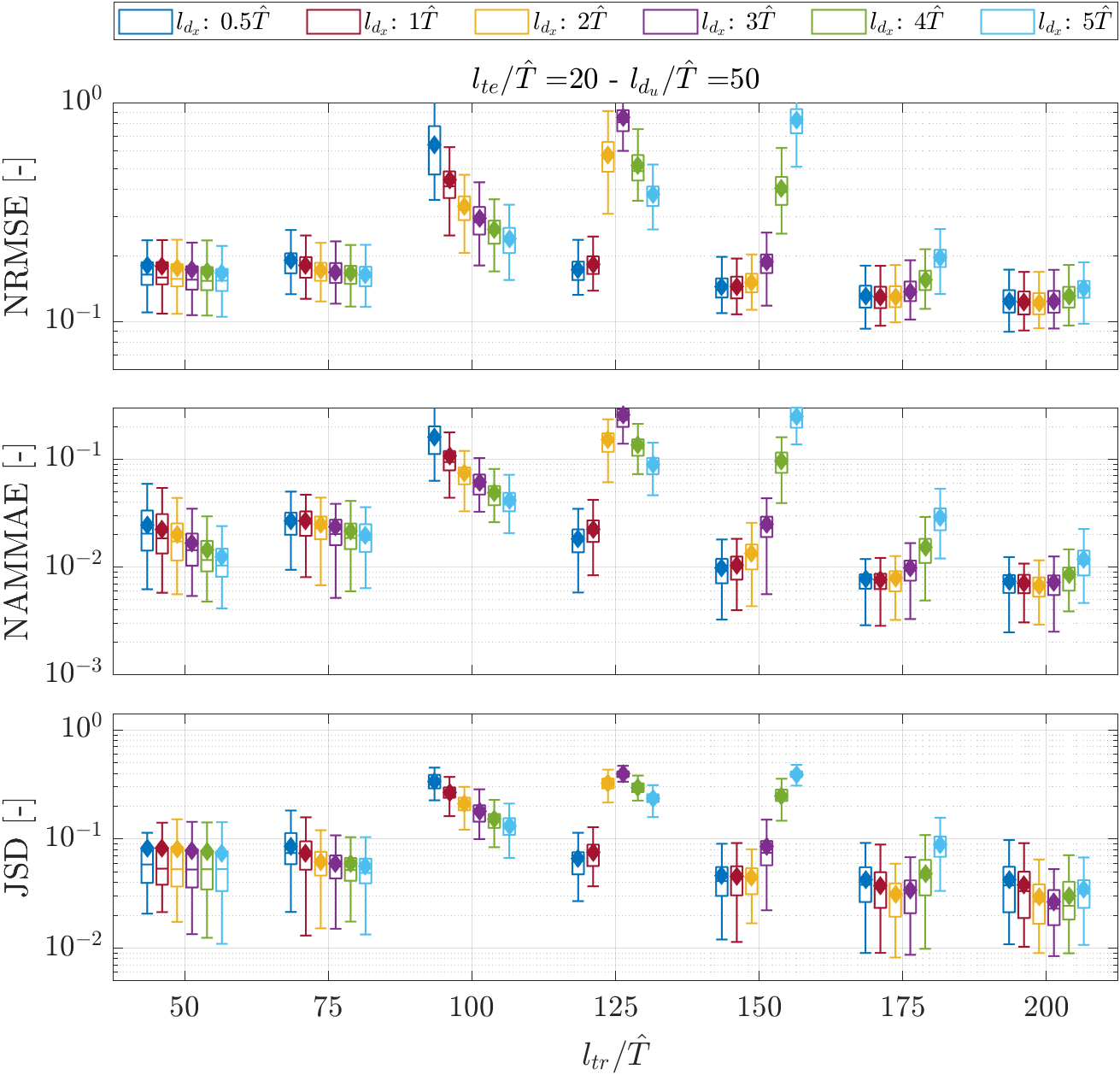}
    \caption{Hankel-DMDc, boxplot of error metrics over the validation set for tested $l_{tr}$ and $l_{d_x}$ with $l_{d_u}/\hat{T}=50$. Diamonds indicate the average value of the respective configuration. $l_{te}=20\hat{T}$ ($\sim 146s$).}
    \label{fig:simet6}
\end{figure}

Results from the design-of-experiment on system identification via deterministic Hankel-DMDc are reported in terms of boxplots graphs in
\cref{fig:simet1,fig:simet2,fig:simet3,fig:simet4,fig:simet5,fig:simet6} combining the outcomes of the 100 training/validation sequences combinations for each hyperparameter configuration. 
The metrics are evaluated using a prediction timeframe of $l_{te}=20\hat T$ ($\sim 146s$).
\begin{figure}
    \centering
    \includegraphics[width=\linewidth]{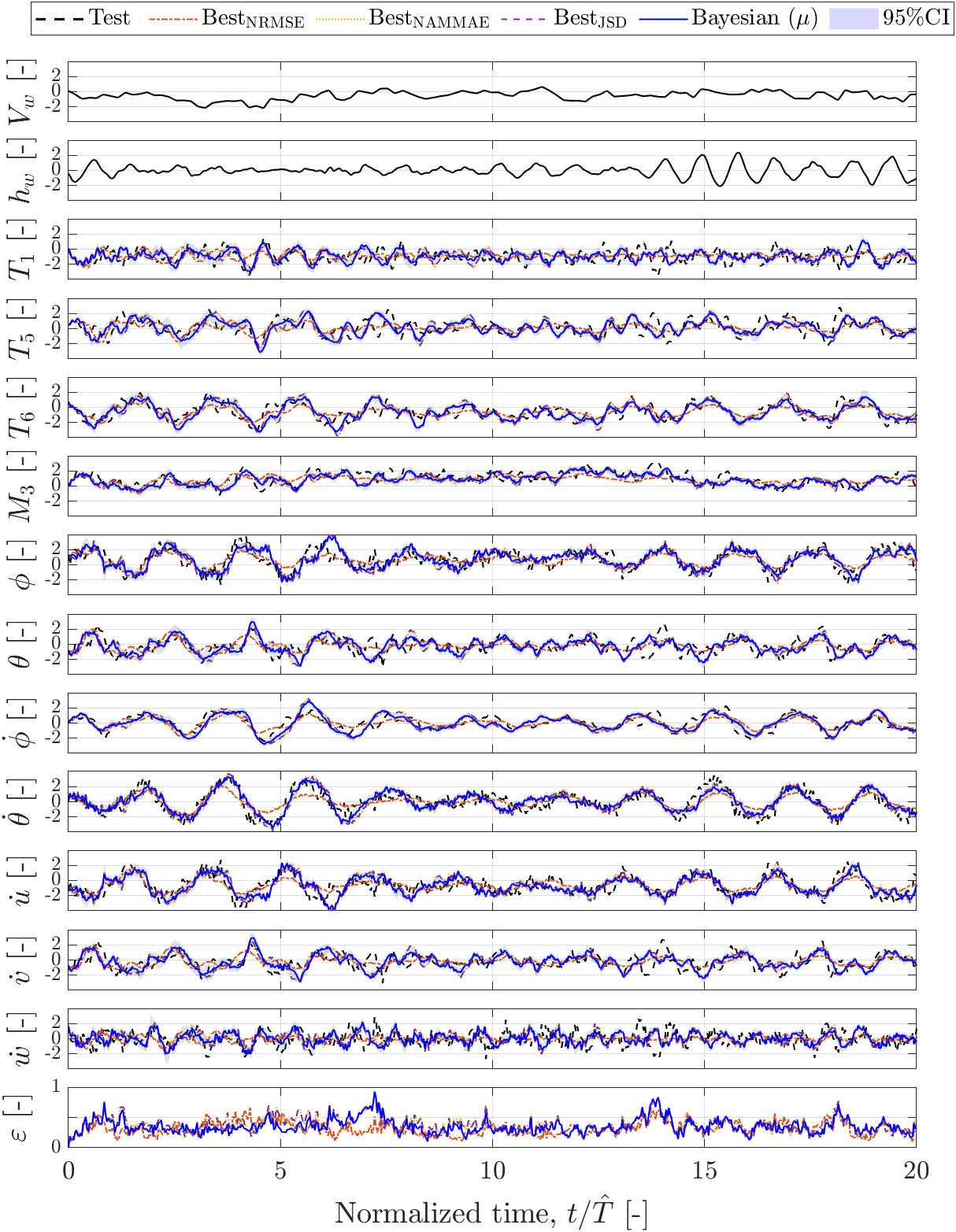}
    \caption{Standardized time series prediction by deterministic (hyperparameters for best average metrics) and Bayesian Hankel-DMDc. Selected sequence 1.}
    \label{fig:si6}
\end{figure}
\begin{figure}
    \centering
    \includegraphics[width=\linewidth]{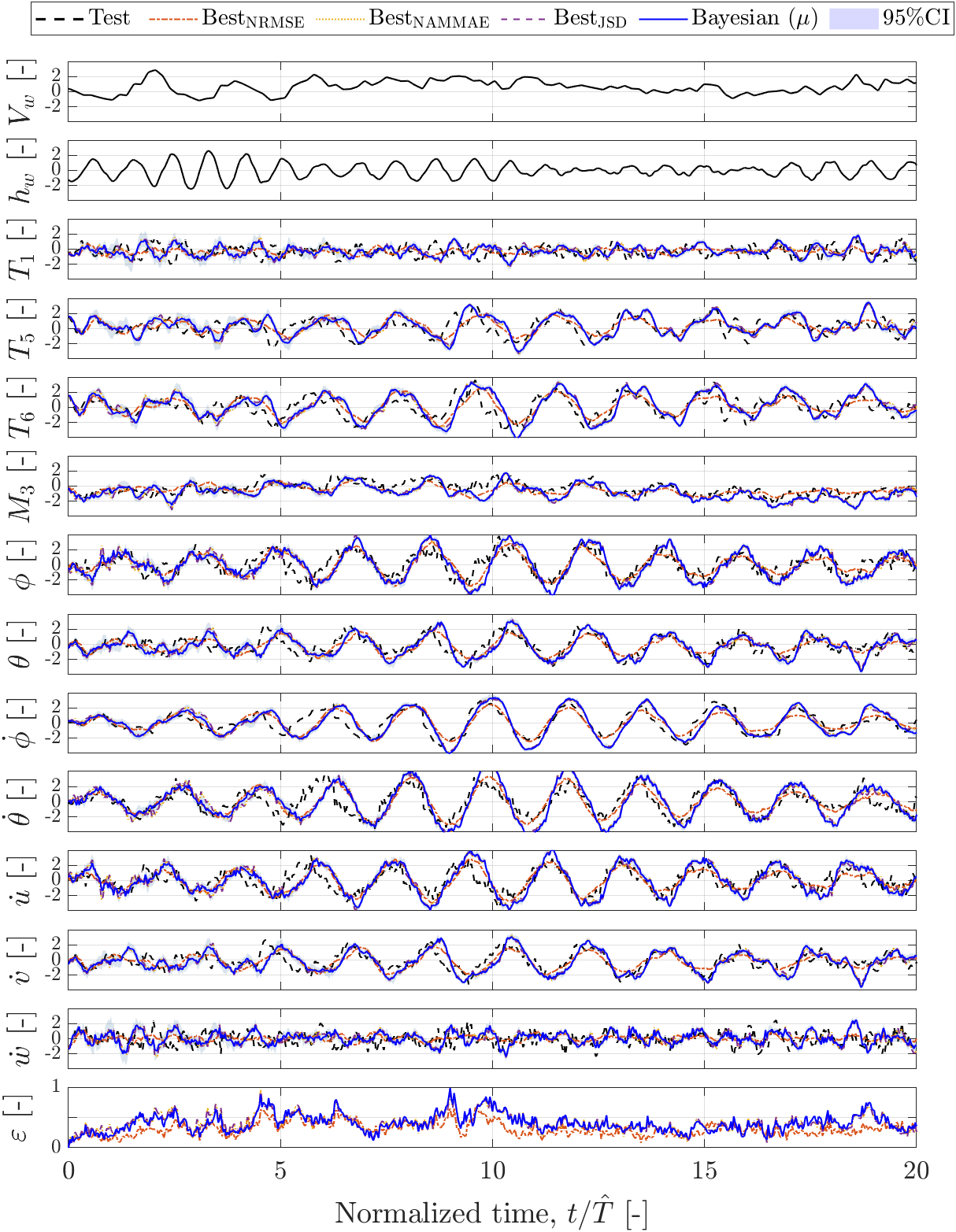}
    \caption{Standardized time series prediction by deterministic (hyperparameters for best average metrics) and Bayesian Hankel-DMDc. Selected sequence 2.}
    \label{fig:si9}
\end{figure}
\begin{figure}
    \centering
    \includegraphics[width=\linewidth]{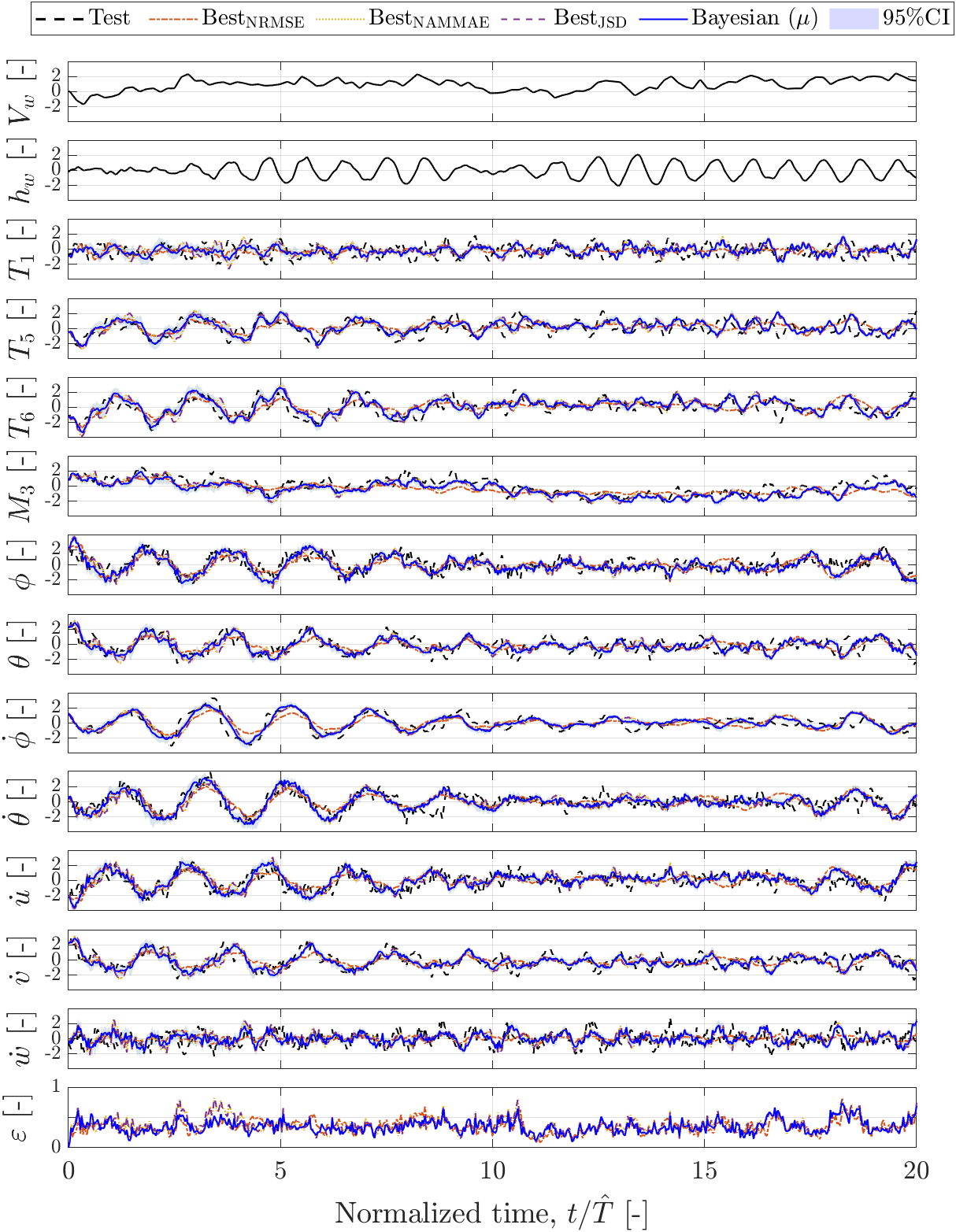}
    \caption{Standardized time series prediction by deterministic (hyperparameters for best average metrics) and Bayesian Hankel-DMDc. Selected sequence 3.}
    \label{fig:si10}
\end{figure}
\begin{figure}
    \centering
    \includegraphics[width=\linewidth]{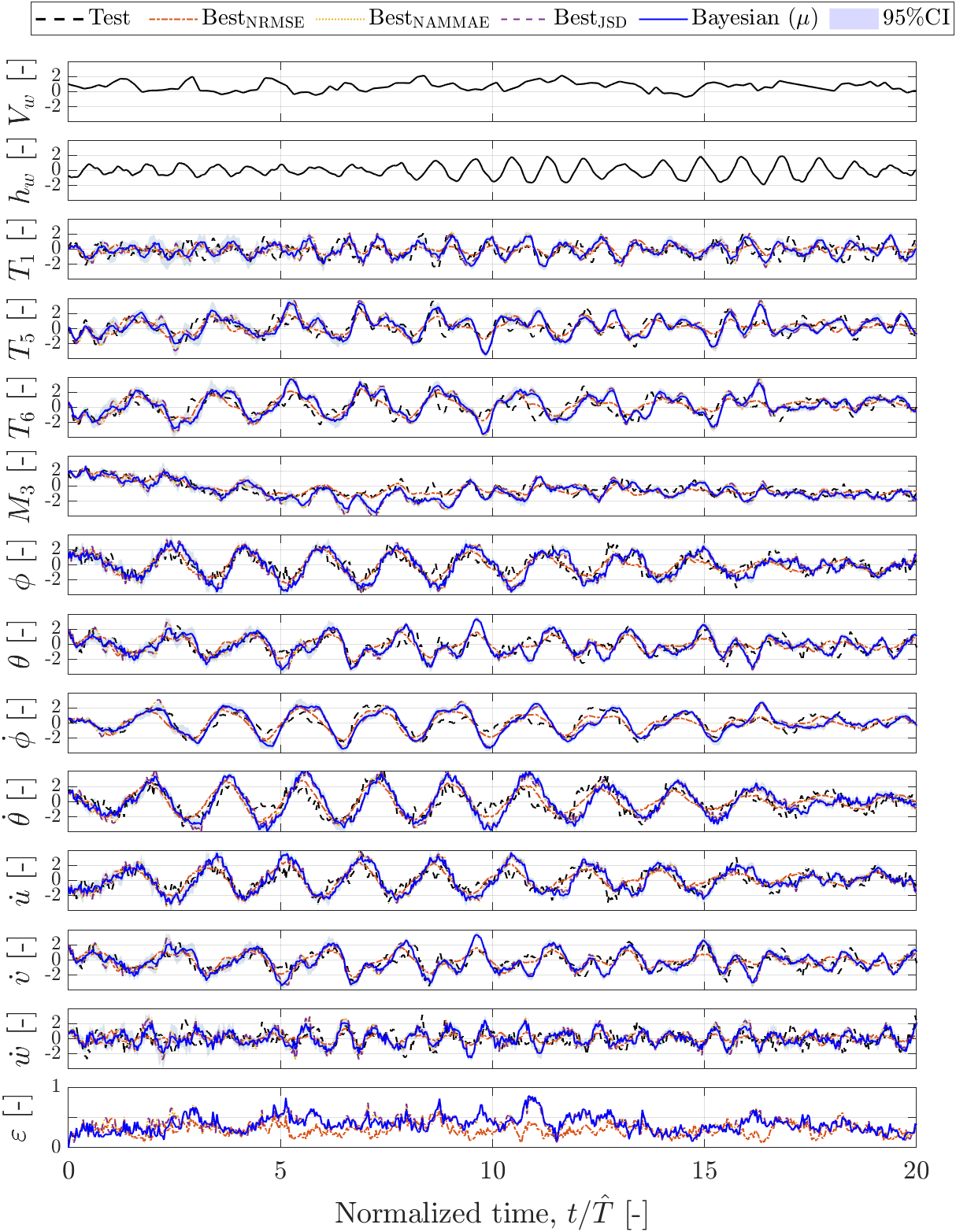}
    \caption{Standardized time series prediction by deterministic (hyperparameters for best average metrics) and Bayesian Hankel-DMDc. Selected sequence 4.}
    \label{fig:si20}
\end{figure}

Analyzing the boxplots, some configurations of the hyperparameters are characterized by very high values of all the metrics. 
The $\mathbf{A}$ matrices of the pertaining Hankel-DMDc models have been noted to have eigenvalues with positive real parts, causing the predictions to be unstable. 
As pointed out by \cite{rains2024}, DMDc algorithms are particularly susceptible to the choice of the model dimensions and prone to identify spurious unstable eigenvalues. 
In the nowcasting approach with Hankel-DMD, this phenomenon can be effectively mitigated by projecting the unstable discrete-time eigenvalues onto the unit circle. However, in system identification with Hankel-DMDc, the $\mathbf{B}$ matrix is also affected by the identification of unstable eigenvalues, and no simple stabilization procedure is available.  

\Cref{tab:detbestconfsi} reports the value of the hyperparameters for the configurations producing the best average value of NRMSE, NAMME, and JSD. 
It emerges that the most accurate predictions are obtained with high values of training length, $l_{tr}<175\hat{T}$. 
The noisy nature of the experimental measures and the variability of operational conditions encountered by the FOWT during the acquisitions can partially explain this result.
A large $l_{tr}$ implies a training signal that extends over a broader set of operative situations, allowing capturing the system's relevant features with increased effectiveness.
A subdomain of the explored range for the hyperparameters value, defined as $175\hat{T}\le l_{tr} \le 200\hat{T}$, $2\hat{T}\le l_{d_x} \le 4\hat{T}$, and $20\hat{T}\le l_{d_u} \le 40\hat{T}$, includes configurations with low values of all the there metrics.
The Best$_\text{NAMMAE}$ and Best$_\text{JSD}$ configurations are similar and lie in this range for which also the NRMSE shows small values (although not the best one). On the contrary, the Best$_\text{NRMSE}$ setup is characterized by a small value of $l_{d_u}$, in a range that shows higher NAMMAE and JSD errors. 

\Cref{fig:si6,fig:si9,fig:si10,fig:si20} show the predictions by the Hankel-DMDc for random test sequences taken as representative. 
The figures show the input variables with a solid black line, the test sequence in a dashed black line, and the prediction obtained with the Best$_\text{NRMSE}$, Best$_\text{NAMMAE}$ and Best$_\text{JSD}$ hyperparameters, as reported in \cref{tab:detbestconfsi}, with an orange dash-dotted, yellow dotted and purple dashed line respectively.
Best$_\text{NAMMAE}$ and Best$_\text{JSD}$ produce similar predictions, following their similarity in hyperparameters values. 
The Best$_\text{NRMSE}$ line behaves slightly differently from the other two and seems to capture less high-frequency oscillations of the test sequence.

As a general consideration, even though a phase-resolved agreement is not obtained, the predictions show a strong statistical similarity with the test signals, as testified both qualitatively by \cref{fig:si6,fig:si9,fig:si10,fig:si20} and quantitatively by the JSD boxplots.

It can be noted that, for all the plotted configurations, the value of $\varepsilon$ does not show a monotonically increasing trend as prediction time progresses, or in other words, the prediction accuracy is not negatively proportional to the prediction horizon. This suggests that the systems identified by the Hankel-DMDc may keep the shown level of accuracy indefinitely in time. This characteristic is fundamental for the proposed usages of the method.

\subsection{System identification via Bayesian Hankel-DMDc}\label{s:sibhdmdc}
As done for the nowcasting, the Bayesian extension of the Hankel-DMDc algorithm for system identification is obtained by exploiting the insights on the hyperparameters derived from the deterministic analysis, in particular for the identification of a suitable range of variation of $l_{tr}$, $l_{d_x}$, and $l_{d_u}$. 

The three hyperparameters are treated as probabilistic variables, uniformly distributed in $l_{tr}/{\hat{T}}$~$\sim$~$\mathcal{U}(175,200)$, $l_{d_x}/{\hat{T}}$~$\sim$~$\mathcal{U}(2,4)$, and $l_{d_u}/{\hat{T}}$~$\sim$~$\mathcal{U}(20,40)$ (the actual $n_{d_x}$ and $n_{d_u}$ are taken as the nearest integers from the calculated values).

\Cref{fig:si6,fig:si9,fig:si10,fig:si20} show a comparison between the prediction by the Bayesian system identification and the deterministic best configurations, with the blue shadowed area representing the 95\% confidence interval of the stochastic prediction.
As observed for the deterministic analysis, no accuracy degradation trend with the prediction time is noted for the Bayesian prediction.
The Bayesian expectation is very close to the Best$_\text{NAMMAE}$ and Best$_\text{JSD}$, with small uncertainty. This reflects a general \textit{robustness} of the deterministic predictions in the range of variation of the stochastic parameters, already noted in the similarity between the  Best$_\text{NAMMAE}$ and Best$_\text{JSD}$ solutions.

\begin{figure}
    \centering
\includegraphics[width=0.5\linewidth]{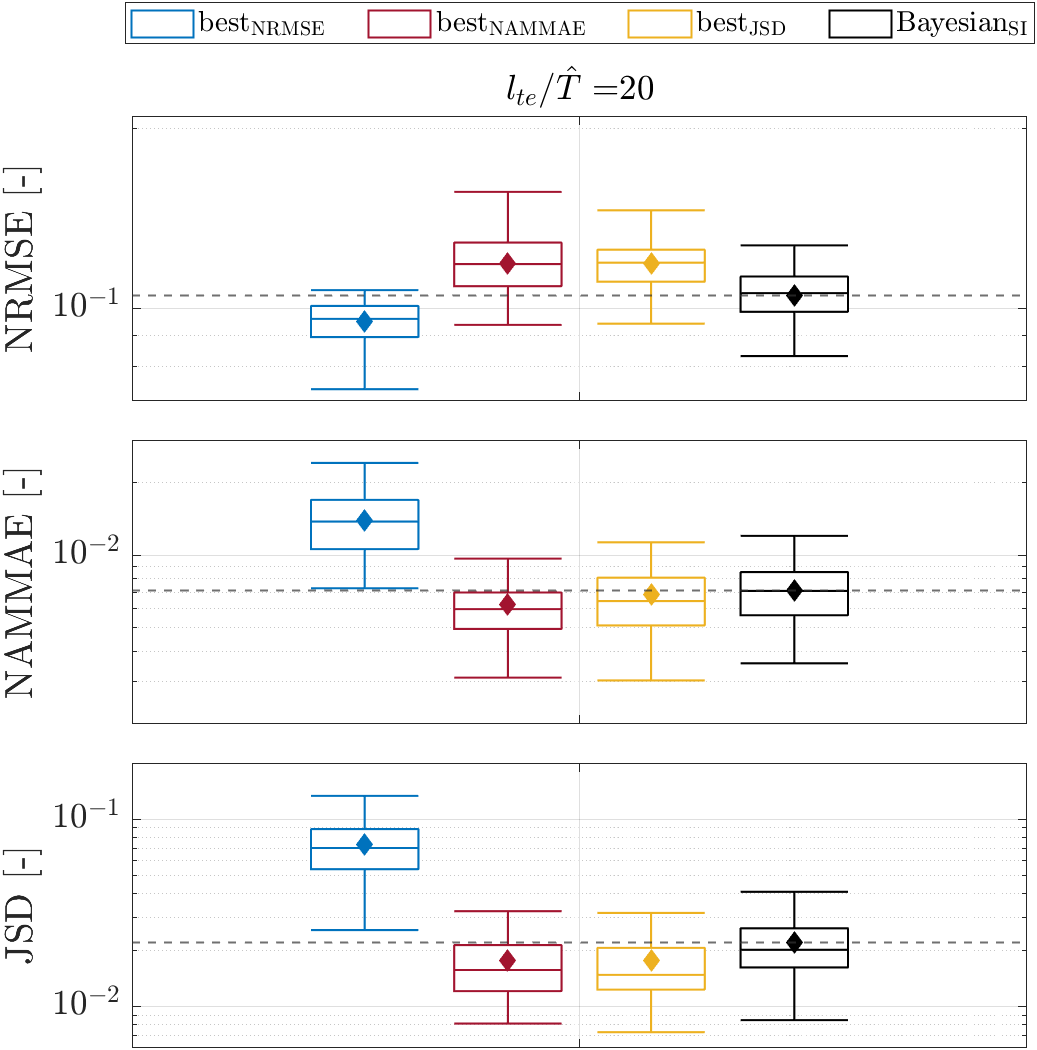}
    \caption{Error metrics comparison, Hankel-DMDc vs. Bayesian Hankel-DMDc system identification, $l_{te}=20\hat T$ ($\sim 146s$.}
    \label{fig:sidetbay}
\end{figure}

A comparison between the best deterministic configurations and the Bayesian system identification in terms of NRMSE, NAMMAE, and JSD metrics is presented in \cref{fig:sidetbay} for $l_{te}=20\hat{T}$. 
The Bayesian approach improves the NRMSE compared to the deterministic solutions, preserving the NAMMAE and JSD results that are only slightly degraded.
\begin{table}
\caption{Resume of the best hyperparameters configurations for system identification as identified by the deterministic design of experiment, $l_{te}=20 \hat{T}$, and Bayesian setup.}\label{tab:detbestconfsi}
\begin{tabular}{lllllll}
\toprule
                                   & NRMSE & NAMMAE & JSD   & $l_{tr}$       & $l_{d_x}$      & $l_{d_u}$     \\
                                   & (avg) &  (avg) & (avg) &                &                &               \\
      \midrule
       $\text{Best}_\text{NRMSE}$  &0.118 &0.022 &0.073 & 200$\hat{T}$     & 2$\hat{T}$     & 5$\hat{T}$    \\
       $\text{Best}_\text{NAMMAE}$ &0.149 &0.010 &0.018 & 200$\hat{T}$     & 3$\hat{T}$     & 40$\hat{T}$    \\
       $\text{Best}_\text{JSD}$    &0.149 &0.011 &0.017 & 175$\hat{T}$     & 4$\hat{T}$     & 30$\hat{T}$    \\       
       Bayesian                    &0.131 &0.011 &0.021 & [175-200]$\hat{T}$ & [2-4]$\hat{T}$ & [20-40]$\hat{T}$\\
     \midrule
\end{tabular}
\end{table}

\begin{figure}
    \centering
    \includegraphics[width=\linewidth]{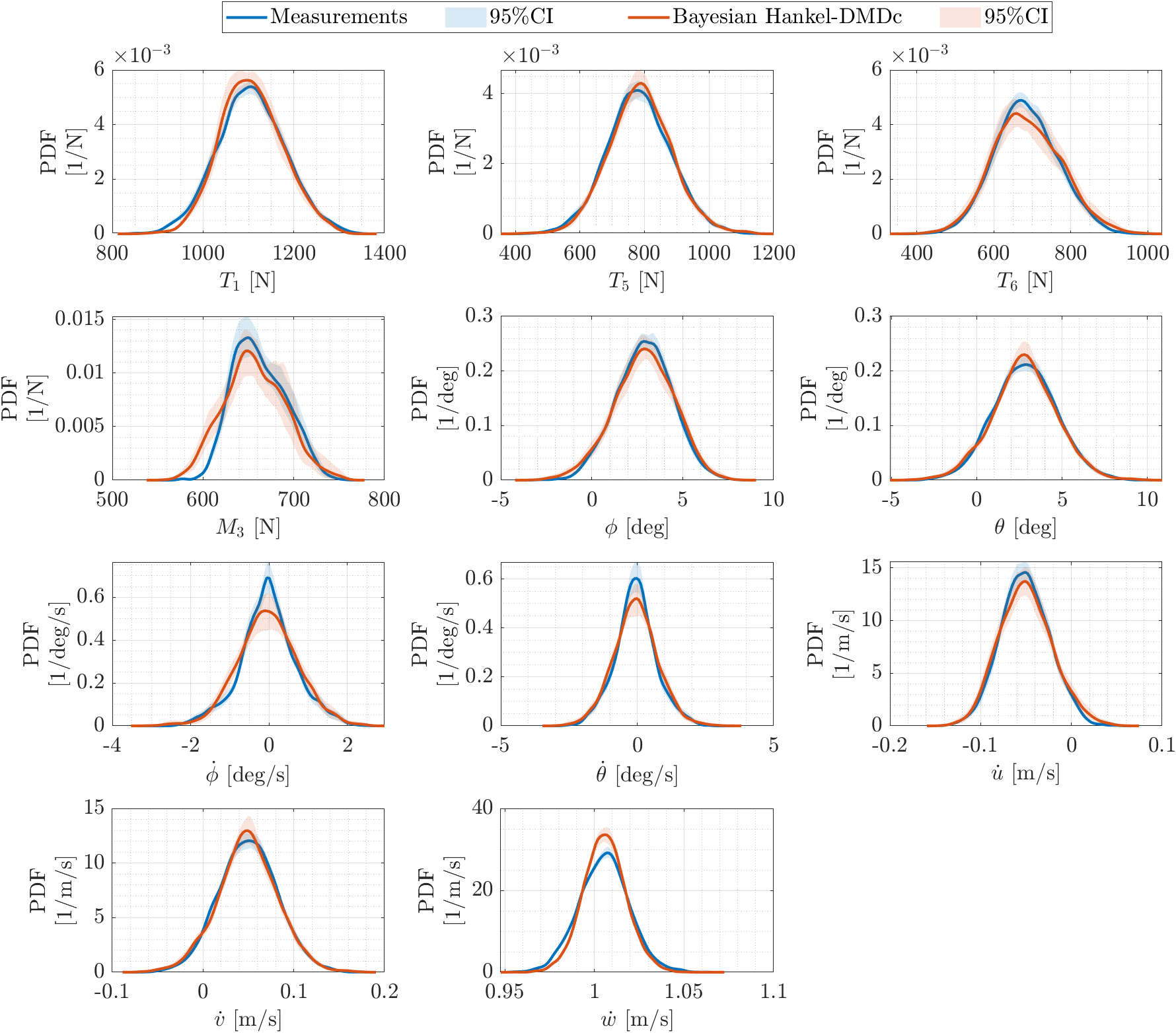}
    \caption{Probability density function comparison between measured data and the expected value of the Bayesian Hankel-DMDc prediction on bootstrapped sequences. Shaded areas indicate the 95\% confidence interval of the two PDFs.}
    \label{fig:sipdf}
\end{figure}

The effectiveness of the Bayesian Hankel-DMDc ROM as a surrogate model is further assessed by statistically comparing the probability density functions (PDFs) of the time series generated by the Bayesian system identification and the experimental measurements. 
A moving block bootstrap (MBB) method is employed to define the time series for analysis following \cite{serani2021urans}, which introduces uncertainty in the PDFs estimation using 100 bootstrapped series. 
The PDF of each time series is calculated from the expected value of the predictions obtained by the ROM and the original data using kernel density estimation \cite{Miecznikowski2010} as follows:
\begin{equation}
    \text{PDF}\left(x,y\right) = \frac{1}{\mathcal{T} h}\sum_{i=1}^{\mathcal{T}} K \left(\frac{y -x_i}{h}\right).
\end{equation}
Here, $K$ is a normal kernel function defined as
\begin{equation}
    K\left(\xi\right) = \frac{1}{\sqrt{2 \pi}} \exp{\left(-\frac{\xi^2}{2}\right)},
\end{equation}
where $h=\sigma(x)\mathcal{T}^{-1/5}$ is the bandwidth \cite{Silverman2018}. 
The quantile function $q$ is evaluated at probabilities $p=0.025$ and $0.975$, defining the lower and upper bounds of the 95\% confidence interval as $U_{PDF(\xi,y)} = PDF(\xi,y)_{q=0.95} - PDF(\xi,y)_{q=0.025}$. 
\Cref{fig:sipdf} shows the expected value of the PDFs over the bootstrapped series as solid lines, with the 95\% confidence interval shown as shaded areas.
\begin{table}[ht!]
    \centering
    \caption{Expected value and 95\% confidence lower bound, upper bound, and interval of JSD of bootstrapped time series}
    \label{tab:mbbjsd}    
    \begin{tabular}{lllll}
    \toprule
             & JSD($\xi$) & \\
    $\xi$    & EV  & q=0.025& q=0.975 & U \\
             \midrule
    $T_1$    & 0.0039 & 0.0018 & 0.0072 & 0.0054\\
    $T_5$    & 0.0021 & 0.0009 & 0.0043 & 0.0034\\
    $T_6$    & 0.0042 & 0.0011 & 0.0107 & 0.0096\\
    $M_3$    & 0.0218 & 0.0100 & 0.0376 & 0.0276\\
    $\phi$   & 0.0021 & 0.0009 & 0.0041 & 0.0032\\
    $\theta$ & 0.0037 & 0.0009 & 0.0086 & 0.0077\\
$\dot \phi$  & 0.0040 & 0.0011 & 0.0091 & 0.0080\\
$\dot \theta$& 0.0021 & 0.0009 & 0.0040 & 0.0032\\
$\dot u$    & 0.0068 & 0.0036 & 0.0110 & 0.0074\\
$\dot w$    & 0.0079 & 0.0033 & 0.0176 & 0.0144\\
$\dot w$    & 0.0047 & 0.0013 & 0.0113 & 0.0100\\
    \midrule
    avg      & 0.0058 & 0.0024 & 0.0114 & 0.0091\\
    \midrule
    \end{tabular}
\end{table}
Results show a good overall agreement between the distributions obtained from ROM and experimental measurements.
The statistics of large amplitude motions and loads is well captured. The largest differences are observed for the derivative of pitch and roll angles in the small values range.
The confidence intervals of the Bayesian Hankel-DMDc predictions adequately cover the PDF of the measured data, indicating that the predictions are accurate and reliable.

The statistics of the JSD evaluated on the bootstrapped time series for the eleven variables are reported in \cref{tab:mbbjsd} to quantify the differences between the experimental and DMD distributions in \cref{fig:sipdf}. 
The analysis confirms the similarity between the probability distributions of Bayesian Hankel-DMDc and real data sequences.
The small value of the JSD expected values is a sign of statistically accurate and reliable predictions, which is important for ROM application in lifecycle assessment and maintenance planning.

The training time for the system identification is investigated as in \cref{s:ncbhdmd} on the same mid-end laptop, averaging over 10 realizations using 100 different hyperparameter values in the intervals of the Bayesian analysis. The larger sizes of data managed compared to the nowcasting task results in larger matrices and longer times, ranging from $\mu_t=21.512 s$, $\sigma_t=1.493 s$ to $\mu_t=41.116 s$, $\sigma_t=3.027 s$, 
still computationally cheaper compared to high-fidelity simulations or typical training of deep-learning methods, and also more data-lean of the latter ones.

\section{Conclusions} \label{s:conc}
This work explored for the first time, to the best of the authors' knowledge, the use of DMD and some of its algorithmic variants to extract knowledge, perform forecasting, and system identification from real-life measured data of a floating offshore wind turbine.
In this work, the tasks are performed on experimentally measured data, but they also directly apply to other data sources such as simulations of various fidelity levels.

The modal analysis shows the coupling between the floater motion, loads on tendons and moorings, and the elevation of the wave encountered by the platform. Interestingly, the power extracted by the wind turbine and the rpm of its blades are almost not involved in the description of the floating motions and seem scarcely influenced by them.

The Hankel-DMD has been successfully used to obtain a data-driven, equation-free, data-lean, and continuously learning short-term forecasting method, \textit{nowcasting}, for the motion of the wind turbine platform and the loads acting on its tendons and moorings.
The Hankel extension of the DMD allows the method to capture and represent essential features of nonlinear dynamical systems such as the FOWT within a linear system, spanning an augmented coordinate system that includes time-delayed copies of the variables in the system state. 
The nowcasting algorithm proved to predict FOWT motions, accelerations, and mooring/tendons loads effectively up to 4 wave encounter periods, with a computational cost compatible with real-time execution and the needed accuracy for applications such as model predictive control and short-term digital twinning.
A full-factorial design of experiment identified reasonable variation ranges of the Hankel-DMD hyperparameters to obtain accurate results, using three metrics (NRMSE, NAMMAE, and JSD) to assess different aspects of the prediction errors.
A systematic difficulty is encountered in predicting the turbine extracted power and blades rotational speed, which signals expose the largest nonlinear characteristics due to the turbine control system. 
The Bayesian extension of the Hankel-DMD included uncertainty quantification in the analysis considering the hyperparameters of the method as stochastic variables varying in the previously identified ranges, also obtaining accurate results.

The system identification task is performed by applying the Hankel-DMDc method, which models the FOWT dynamical system as externally forced. In particular, the considered input vector includes the wave elevation measured by the ADCP about 50 m SE from the FOWT and the wind speed measured by the PLC placed on the turbine nacelle.
The inclusion of forcing terms in the reduced order model permits the extension of the prediction to a longer time horizon, providing the knowledge at each prediction time step of the current value of the input variables. 
Once trained, the resulting data-driven models are suitable to be used in place of the original FOWT model, providing statistically accurate predictions of the modeled variables with a reduced computational cost and possibly undefined time extension, with possible useful applications in control, life-cycle assessment, maintenance, and operational planning.
Hankel-DMDc is also extended to a Bayesian version, with the same rationale used for the Hankel-DMD, obtaining the same benefits.

The noisiness of the measured data and the strong nonlinearities, which also arise from the extreme weather conditions of this specific dataset, notoriously constitute a demanding challenge for methods based on DMD. Nevertheless, the results obtained are promising from both nowcasting and system identification. 
Future works may explore the possibility of hybridizing the Hankel-DMD-based methods with some machine learning techniques, to further improve its noise rejection properties and its capability of capturing nonlinear features of the system at hand, trying to preserve the peculiarity of the original method.

A challenge for the system identification task in the current study is the complexity and variety of operational conditions faced by the FOWT during the field measurements.
Such a complete dataset in terms of system forcing and operational conditions is, on one side, precious to test the DMD-based system identification in a realistic environment.
On the other side, this requires the method to use a long training signal and a large number of additional delayed variables in the strive to identify a comprehensive model, which is in any case a non-trivial task.
Optimizing the training set and the characteristics of the data collected may represent an important development to improve the model's accuracy.
A different approach may also be tested: firstly applying the system identification to data collected in more specific conditions/excitations (such as single wave headings, specific magnitude ranges and direction of wind, etc.) and secondly using an interpolation of parametric reduced-order models \cite{Farhat2008,Farhat2011} for generalizing the models.

In addition, the current experimental setup does not allow the identification of the incoming wave direction, as it includes a single point of measure for $h_w$, nor the wind direction but only their magnitude.
An evolved setup involving multiple wave elevation measurements surrounding the platform, and additional load cells on the other tendons and moorings would provide more complete data that may help in particular for the system identification task.

\section*{Author contributions}
\textbf{Giorgio Palma:} Methodology, Software, Investigation, Validation, Formal analysis, Writing - Original Draft,
Visualization. 
\textbf{Andrea Bardazzi:} Investigation, Data curation, Resources. 
\textbf{Alessia Lucarelli:} Investigation, Data
curation, Resources. 
\textbf{Chiara Pilloton:} Investigation, Data curation, Resources. 
\textbf{Andrea Serani:} Methodology, Writing
- Review \& Editing. 
\textbf{Claudio Lugni:} Investigation, Data curation, Resources, Project administration, Funding
acquisition. 
\textbf{Matteo Diez:} Conceptualization, Methodology, Resources, Writing - Review \& Editing, Supervision.

\section*{Data availability}
Data will be made available upon request.

\section*{Acknowledgements}
This research is supported by the Italian Ministry of the Environment and Energy Security (Ministero dell'Ambiente e della Sicurezza Energetica, MASE) through the Three-Year Plan for Electric System Research (Piano Triennale Ricerca di Sistema Elettrico) 2022-2024.

\bibliographystyle{unsrt}  
\bibliography{cas-refs}  

\end{document}